\documentclass[10pt,twocolumn,letterpaper]{article}

\usepackage{cvpr}              % To produce the CAMERA-READY version
\usepackage[accsupp]{axessibility}
\usepackage{graphicx}
\usepackage{amsmath}
\usepackage{amssymb}
\usepackage{booktabs}
\makeatletter
\@namedef{ver@everyshi.sty}{}
\makeatother
\usepackage{tikz}
\usepackage{multirow}

\usepackage{pifont}
\usepackage{color}
\usepackage{threeparttable}
\usepackage{balance}

\usepackage{pgfplots} % 画折线图
\pgfplotsset{compat=1.7}

\usepackage{scalefnt} %用于设置字体

\definecolor{color1}{RGB}{145,30,180}
\definecolor{color2}{RGB}{245,130,48}
\definecolor{color3}{RGB}{230,25,75}
\usepackage[pagebackref,breaklinks,colorlinks]{hyperref}

\usepackage[capitalize]{cleveref}
\crefname{section}{Sec.}{Secs.}
\Crefname{section}{Section}{Sections}
\Crefname{table}{Table}{Tables}
\crefname{table}{Tab.}{Tabs.}

\def\cvprPaperID{7479} % *** Enter the CVPR Paper ID here
\def\confName{CVPR}
\def\confYear{2023}

\begin{document}

%%%%%%%%% TITLE - PLEASE UPDATE
% \title{Gait Recognition in the 3D World}
% \title{LIDAR GAIT: Benchmarking 3D Gait Recognition with Point Clouds}
% \title{SUSTech1K: Benchmarking 3D Gait Recognition with Point Clouds}
\title{LidarGait: Benchmarking 3D Gait Recognition with Point Clouds}

\author{
Chuanfu Shen$^{1,2}$, 
Fan Chao$^{2,3}$,
Wei Wu$^2$,
Rui Wang$^2$,
George Q. Huang$^{4}$, 
Shiqi Yu$^{2,3}$\thanks{Corresponding Author} \\
{\normalsize $^1$ Department of Industrial and Manufacturing Systems Engineering, The University of Hong Kong} \\
{\normalsize $^2$ Department of Computer Science and Engineering, Southern University of Science and Technology}\\
{\normalsize $^3$ Research Institute of Trustworthy Autonomous System, Southern University of Science and Technology} \\
{\normalsize $^4$ Department of Industrial and Systems Engineering, The Hong Kong Polytechnic University} \\
{\tt\small noahshen@connect.hku.hk, \{12131100, 12032501, 12232385\}@mail.sustech.edu.cn} \\ 
{\tt\small gq.huang@polyu.edu.hk, yusq@sustech.edu.cn}.
}
\maketitle

%%%%%%%%% ABSTRACT
\begin{abstract}
    Video-based gait recognition has achieved impressive results in constrained scenarios. However, visual cameras neglect human 3D structure information, which limits the feasibility of gait recognition in the 3D wild world.
    Instead of extracting gait features from images, this work explores precise 3D gait features from point clouds and proposes a simple yet efficient 3D gait recognition framework, termed \textbf{LidarGait}. Our proposed approach projects sparse point clouds into depth maps to learn the representations with 3D geometry information, which outperforms existing point-wise and camera-based methods by a significant margin.
    Due to the lack of point cloud datasets, we build the first large-scale LiDAR-based gait recognition dataset, \textbf{SUSTech1K}, collected by a LiDAR sensor and an RGB camera.
    The dataset contains 25,239 sequences from 1,050 subjects and covers many variations, including visibility, views, occlusions, clothing, carrying, and scenes.
    Extensive experiments show that (1) 3D structure information serves as a significant feature for gait recognition. (2) LidarGait outperforms existing point-based and silhouette-based methods by a significant margin, while it also offers stable cross-view results. (3) The LiDAR sensor is superior to the RGB camera for gait recognition in the outdoor environment.
    The source code and dataset have been made available at \url{https://lidargait.github.io}.

\end{abstract}

\begin{figure}[ht]

\begin{subfigure}{1\linewidth}
\centering
\includegraphics[width=0.80\linewidth]{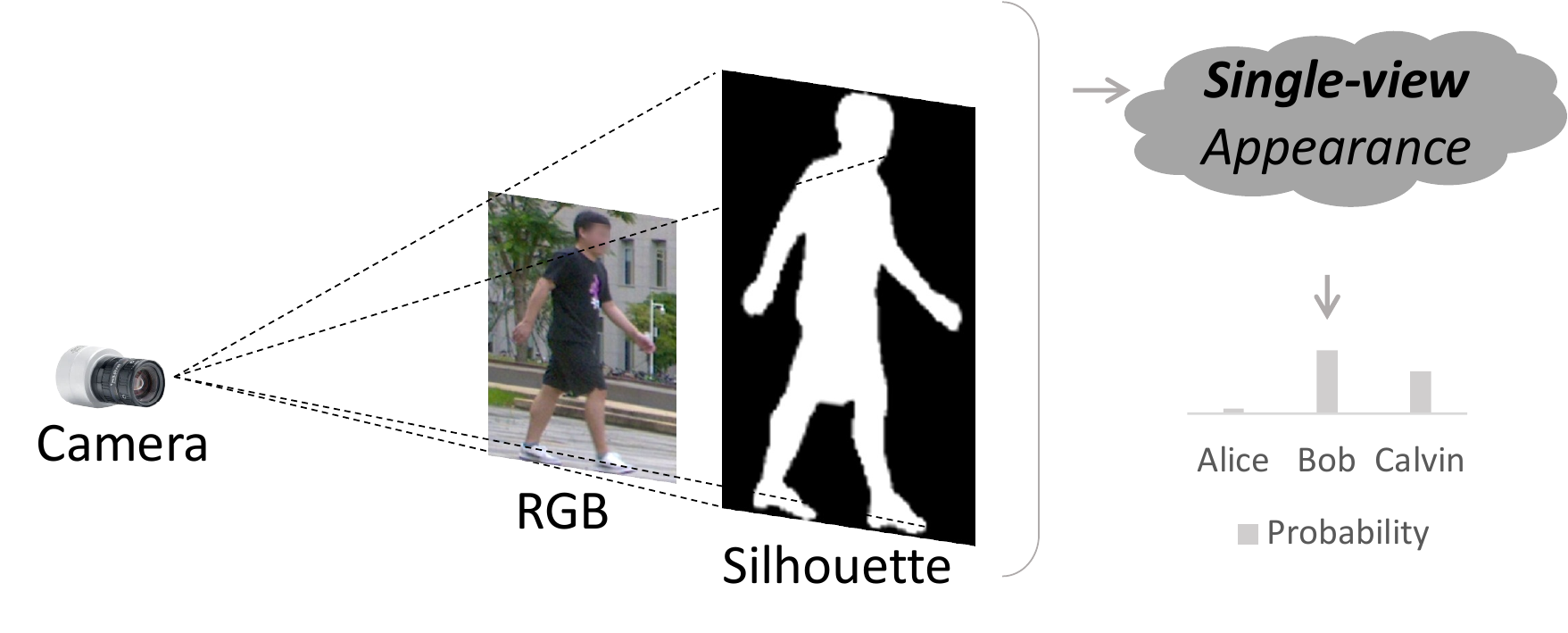}
\caption{Camera-based gait recognition with silhouettes}
\label{fig:silsmethod}
\end{subfigure}

\begin{subfigure}{1\linewidth}
\centering
\includegraphics[width=0.8\linewidth]{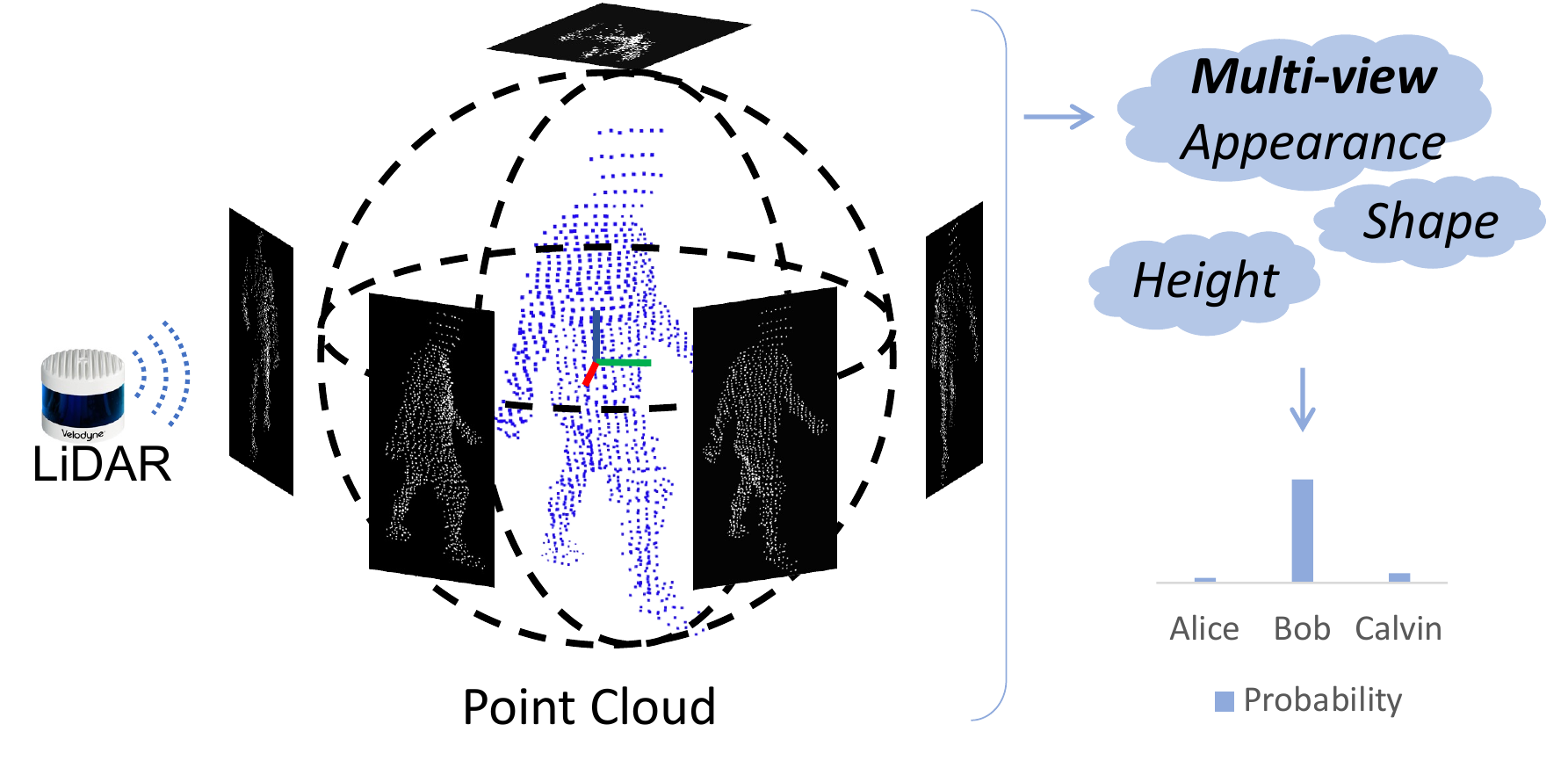}
\caption{LiDAR-based gait recognition with point clouds}
\label{fig:lidarmethod}
\end{subfigure}

\caption{Illustration of \textbf{(a)} camera-based and \textbf{(b)} LiDAR-based gait recognition. Camera-based gait recognition commonly uses silhouettes to learn shape information from a single view. LiDAR-based gait recognition can use 3D structure, shape, and scale information to identify a subject.}
\vspace{-3mm}
% \vspace{-1.5em}
\end{figure}

%%%%%%%%% BODY TEXT
\section{Introduction}
\label{sec:intro}

Gait is an essential biometric, which has the unique advantage of human identification at a distance without physical contact. Gait empowers real-world applications such as human retrieval, forensic identification, and serving robots. 
Recently, great progress has been made to promote gait recognition from in-the-lab setting~\cite{casiab,oulp,oumvlp} to in-the-wild scenario~\cite{fvg,gait3d,grew,tumgait}. Despite these studies have made significant contributions to recent advances~\cite{gaitset,gaitpart,doumetagait,gaitgl,hopgait,reversemask,mt3d}, two inherent problems still remain (1) \textit{lack of 3D geometry information}, and (2) \textit{poor feasibility in the real-world scenario}.

Existing camera-based methods~\cite{cstl,gaitnet} are counterintuitive to human nature.  When recognizing a subject~\cite{gait3d,3dreid}, humans consider not only the 2D appearance characteristics, but also 3D geometry structure information like height, shape, and viewpoints. Differently, camera-based gait recognition methods~\cite{liang2022gaitedge,gaitset,gaitgl} either capture 2D representations from a single viewpoint, as shown in Fig.~\ref{fig:silsmethod}, or exploit 3D representations from estimated 3D pose/mesh models~\cite{posegait,lixiangpose,gait3d}, which is usually imprecise in various challenging conditions of low resolution, poor illumination, untrained posture, etc.
Fortunately, 3D sensors provide precise 3D perception like human nature, \eg recognizing a subject from multiple views as illustrated in Fig.~\ref{fig:lidarmethod}.

Visual ambiguity is the alternative limitation of camera-based approaches.
To our knowledge, most existing gait datasets~\cite{oumvlp,grew,gait3d} only consider camera-based modalities, and fail to acknowledge the challenges of visual ambiguity caused by poor illumination and complex backgrounds in outdoor environments. 
These factors can significantly harm the performance of upstream tasks like pedestrian detection and segmentation, which in turn affects the accuracy of the gait system in real-world applications. Thus, obtaining precise 3D information for gait description is highly desirable to eliminate visual ambiguity in RGB images.

The remarkable success of 3D applications~\cite{waymo,kitti,stcrowd} motivates us to endow gait recognition with precise 3D structural information and accurate human perception, by utilizing LiDAR sensors in challenging outdoor environments. In addition to improving gait recognition, LiDAR sensors offer potential benefits in many scenarios, including robotics, healthcare, social security, and surveillance. For example, robots equipped with LiDAR-based gait recognition can function as $24\times7$ security guards, enhancing community safety. Vehicles fitted with LiDAR sensors can aid in locating lost orders and children. Furthermore, LiDAR is more privacy-preserving than cameras, making it suitable for sensitive  scenarios such as nursing homes and kindergartens. Additionally, LiDAR has the potential to enhance biometric security by protecting against Deepfake attacks compared to cameras.

This paper introduces SUSTech1K, the first large-scale LiDAR-based gait dataset to facilitate 3D gait recognition with point clouds. The dataset is captured outdoors using a Velodyne VLS128 LiDAR sensor and an RGB camera mounted together on a robot. Compared to existing datasets listed in Tab.~\ref{tab:dataset}, SUSTech1K offers several distinctive features: (1) \textbf{Precision}. The SUSTech1K dataset provides 3D point clouds as gait representations with high precision and density, providing precise and robust 3D structure information for recognition. (2) \textbf{Scalability}. The dataset captures 25,239 sequences from 1,050 subjects, providing scalability for statistical evaluation. (3) \textbf{Diversity}. The dataset includes diverse and realistic challenges, such as illumination, occlusion, dressing, carrying, and more, along with detailed annotations, enabling the community to study the impact of different factors on gait recognition. (4) \textbf{Multimodality}. The dataset captures data streams from LiDAR and camera sensors, opening up opportunities for exploring sensor fusion approaches for robust gait recognition.

Given that 3D point clouds are formatted differently from pixels in images and that point-based gait recognition has received little attention, we investigate four cutting-edge methods~\cite{pointnet,pointnet++,pointtransformer,simpleview} from the study of point-based object classification~\cite{pointnet}. However, we observed that all the implemented point-based methods performed sub-optimally when compared to methods using camera-based silhouettes. 
We believe the performance gap is primarily due to the difference in feature granularity of the task.
The aforementioned point-based methods are primarily designed for coarse-grained object classification, focusing more on global context modeling. In contrast, gait recognition requires extracting fine-grained local information to achieve high accuracy.

To address this issue, 
we propose a simple yet effective baseline method named the LidarGait. 
Specifically, LidarGait first projects 3D point clouds into depth images from the LiDAR range view and then employs convolutional networks to extract gait features with 3D structural information from the projection. 
This approach contrasts point-wise methods that learn global context from sparse point clouds with limited local connectivity. 
Using convolutional neural networks on projection, LidarGait can efficiently capture the fine-grained and discriminative gait features from sparse point clouds.
Extensive experiments demonstrate that (1) LidarGait is effective in maintaining 3D structural information for gait recognition, and including 3D information can significantly contributes to performance improvement, (2) point-based gait recognition equipped with a LiDAR sensor performs stably well on various challenges, convincingly demonstrating its practical significance.

To summarize, our main contributions are as follows: 
(1) We carry out one of the first studies of 3D gait recognition with point clouds, bringing precise perception and 3D geometry of humans for better practicality in real-world scenarios. 
(2) We introduce SUSTech1K, the first large-scale LiDAR-based gait recognition benchmark, which includes a range of annotations covering occlusions, viewpoints, carrying, clothing, and distance.
(3) We propose a novel point cloud gait recognition framework, LidarGait, outperforming camera-based methods by a large margin.

% % Table 1
\begin{table*}[ht]
\begin{center}
\caption{Comparison of publicly available datasets for gait recognition. }
% 3D, Multimodal, and Wild indicate whether the dataset contains accurate 3D structure annotations, has multiple modality ,is captured in the wild, respectively.
\label{tab:dataset}
\scalebox{0.83}{
\begin{threeparttable}
\small
% \centering
\vspace{-5mm}
\begin{tabular}{lcccccccc}
\midrule[1.5pt]
Dataset & Year & Subject \#     & Seq \#          & View \#     & Data Type                        & 3D                 & Multimodal               & Outdoor                         \\

\midrule[1.5pt]
CASIA-B~\cite{casiab}             & 2006 & 124            & 13,640          & 11          & RGB, Silhouettes                       & \textcolor{red}{\ding{55}}   & \textcolor{red}{\ding{55}}   & \textcolor{red}{\ding{55}}   \\
CASIA-C~\cite{nightgait}             & 2006 & 153            & 1,530           & 1           & Infrared, Silhouettes                  & \textcolor{red}{\ding{55}} & \textcolor{red}{\ding{55}}   & \textcolor{green}{\ding{51}}   \\
KY4D~\cite{ky4d}                        & 2010 & 42             & 168             & 16          & Silhouettes, RGB, 3D Volumetrics  & \textcolor{green}{\ding{51}} & \textcolor{red}{\ding{55}}   & \textcolor{red}{\ding{55}}   \\
TUM-GAID~\cite{tumgait}                    & 2012 & 305            & 3,370           & 1           & Audio, Video, Depth              & \textcolor{green}{\ding{51}}   & \textcolor{green}{\ding{51}}   & \textcolor{red}{\ding{51}}   \\
SZTAKI-LGA ~\cite{lidar2}        & 2016 & 28         & 11         & 1          & 3D Point Cloud                          & \textcolor{green}{\ding{51}}   & \textcolor{red}{\ding{55}}   & \textcolor{green}{\ding{51}}   \\ 
OU-MVLP~\cite{oumvlp}             & 2018 & 10,307         & 288,596         & 14          & Silhouettes                            & \textcolor{red}{\ding{55}}   & \textcolor{red}{\ding{55}}   & \textcolor{red}{\ding{55}}   \\
FVG ~\cite{fvg}        & 2019 & 226         & 2,856         & 3          & RGB                          & \textcolor{red}{\ding{55}}   & \textcolor{red}{\ding{55}}   & \textcolor{green}{\ding{51}}   \\  %lidar
PCG ~\cite{lidar1}        & 2020 & 30         & 60         & 1          & 3D Point Cloud                          & \textcolor{green}{\ding{51}}   & \textcolor{red}{\ding{55}}   & \textcolor{red}{\ding{55}}   \\  %lidar
GREW~\cite{grew}                & 2021 & 26,345         & 128,671         & 882         & Silhouettes, 2D/3D Skeleton, Flow      & \textcolor{red}{\ding{55}}   & \textcolor{red}{\ding{55}} & \textcolor{red}{\ding{55}}   \\
Gait3D~\cite{gait3d}                      & 2022 & 4,000           & 25,309           & 39          & Silhouettes, 2D/3D Skeleton, 3D Mesh    & \textcolor{green}{\ding{51}} & \textcolor{red}{\ding{55}} & \textcolor{green}{\ding{51}} \\
OUMVLP-Mesh ~\cite{oumvlpmesh}        & 2022 & 10,307         & 288,596         & 14          & 3D Mesh                          & \textcolor{green}{\ding{51}}   & \textcolor{red}{\ding{55}}   & \textcolor{red}{\ding{55}}   \\ \hline
\textbf{SUSTech1K}         & 2023    & \textbf{1,050} & \textbf{25,239} & \textbf{12} & \textbf{RGB, Silhouettes, 3D Point Cloud} & \textcolor{green}{\ding{51}} & \textcolor{green}{\ding{51}} & \textcolor{green}{\ding{51}}\\ 
\end{tabular}
\end{threeparttable} %\vspace{-3mm}
}
\end{center} \vspace{-5mm}
\end{table*}

\section{Related Work}
\label{sec:related}

\noindent\textbf{Gait Recognition.} According to the used representations, gait recognition can be generally divided into 2D and 3D representations-based methods~\cite{survey}.

The majority of 2D representations-based methods study gait characteristics directly from images, termed appearance-based~\cite{casiaa,gaitset,gaitpart,oumvlp} methods, which have made surprising high performance based on silhouettes~\cite{liang2022gaitedge,gei,lixiangpose2,lin2021gaitmask,lin2022uncertainty} together with other gait templates~\cite{gei,wang2010chrono,MEI}. 
The alternative approaches learn human structure~\cite{posegait,gaitgraph,lixiangpose} and dynamics~\cite{gaitgraph} as gait representations, but they are heavily constrained by model-based estimation models.
3D representation-based methods are generally extracted by sensors~\cite{tumgait,kinect} or estimation models~\cite{posegait,lixiangpose2}. The commonly used 3D sensors such as Kinect, provide 3D structured data, but they only facilitate in an indoor and close-distance environment~\cite{kinect}. Meanwhile, multi-cameras reconstruction~\cite{tunnel} and 3D estimation models~\cite{gaitgraph,posegait,gaitnet,gait3d,lixiangpose2} provide considerable 3D geometry, but the performance is far behind the requirements of real-world applications as reported in~\cite{grew}.

\noindent\textbf{Gait Recognition Benchmark.} There are three types of publicly available datasets: in-the-lab~\cite{casiab,oumvlp,oulp}, synthetic~\cite{dou2021versatilegait}, and in-the-wild datasets~\cite{grew,gait3d,resgait,tumgait}. The in-the-lab datasets~\cite{casiaa,casiab,oumvlp,oulp}, represented by CASIA series~\cite{casiaa,casiab,nightgait} and OU-ISIR series~\cite{oumvlp,oulp}, advance the investigation of the feasibility of gait recognition.
The recent synthetic datasets~\cite{dou2021versatilegait} are to overcome the difficulty in data acquisition and annotation of gait, providing more synthetic data with a variety of annotations but introducing cross-domain issue~\cite{liang2022gaitedge} at the same time.
The in-the-wild datasets~\cite{resgait,grew,gait3d} are to promote gait recognition research in the unconstrained environment.
The recent works~\cite{lidar1,lidar2,lidarmethod} based on LiDAR sensor are closely related to our work, while the main concern is that the existing datasets include at most 30 subjects, which cannot guarantee statistically reliable performance evaluation of LiDAR-based gait recognition. Because of insufficient 3D representations for data-driven gait recognition, as shown in Tab.~\ref{tab:dataset}, a dataset with accurate 3D representations is essential.

\noindent\textbf{Point Cloud and 3D Object Classification.} LiDAR, which stands for Light Detection and Ranging, projects laser pulses to the targets and then generate point cloud sets. Each point represents a data point in Cartesian coordinates $(X, Y, Z)$. Point cloud data is sparsely distributed, remaining a significant challenge in modeling correlation and geometry. 3D object classification explore projection-based~\cite{mvcnn,simpleview,viewgcn}, point-wise~\cite{pointnet,pointnet++,pointtransformer}, and graph-wise models~\cite{DGCNN,viewgcn} to capture discriminative feature on point cloud data for object classification. In this paper, we select many representative models of 3D object classification and compare them with our proposed method to comprehensively study 3D point-based gait recognition.

\begin{figure}[t]

\centering
\includegraphics[width=0.8\linewidth]{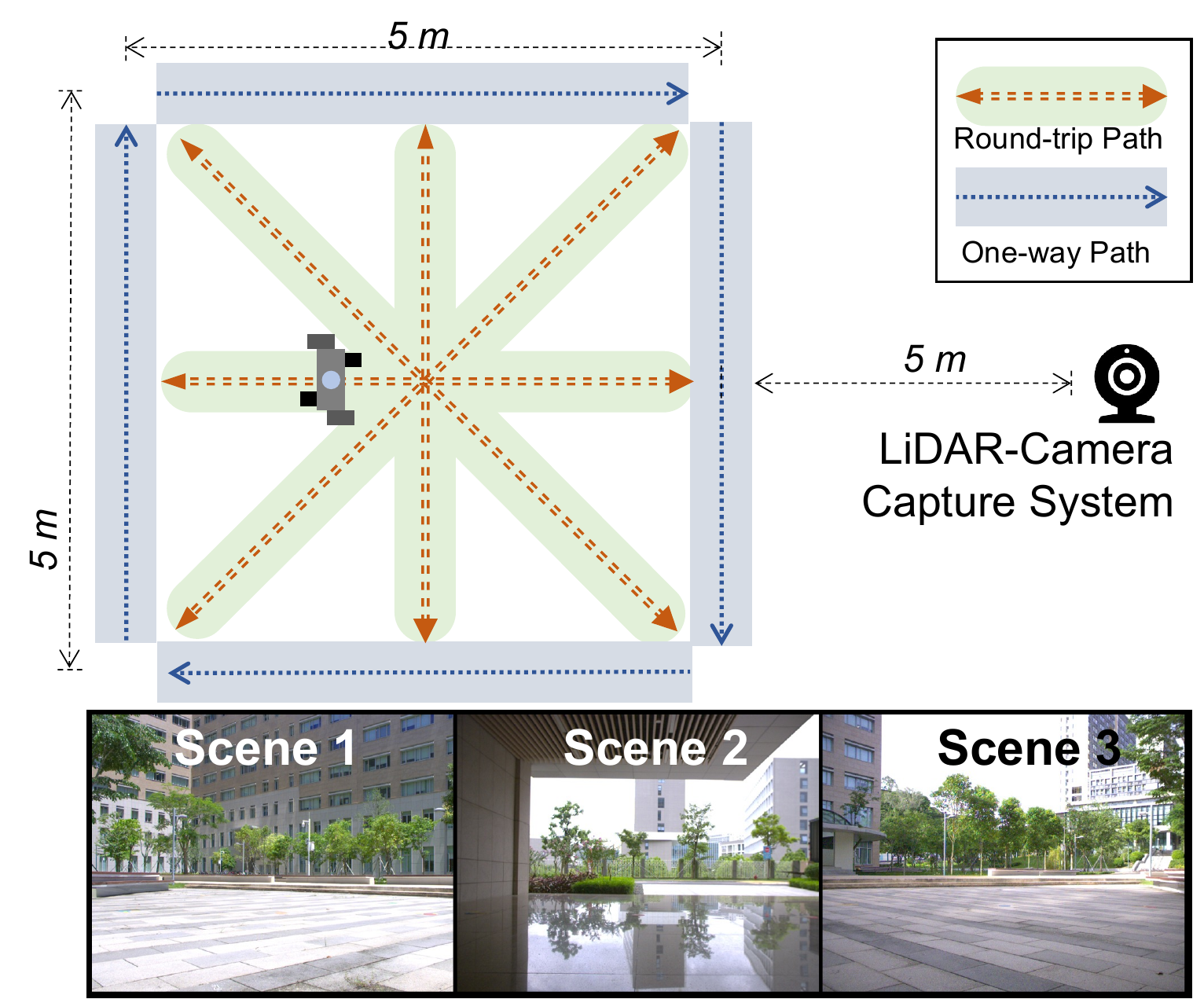}
\caption{Data acquisition setup. Each participant is first instructed to normally walk along four round-trip paths and four one-way paths, then walk again with a random variance along the same paths.}
\label{fig:setup}
\vspace{-5mm}
\end{figure}

\begin{figure*}[t]
     \centering
     \begin{subfigure}[b]{0.72\textwidth}
         \centering
         \includegraphics[width=1\textwidth]{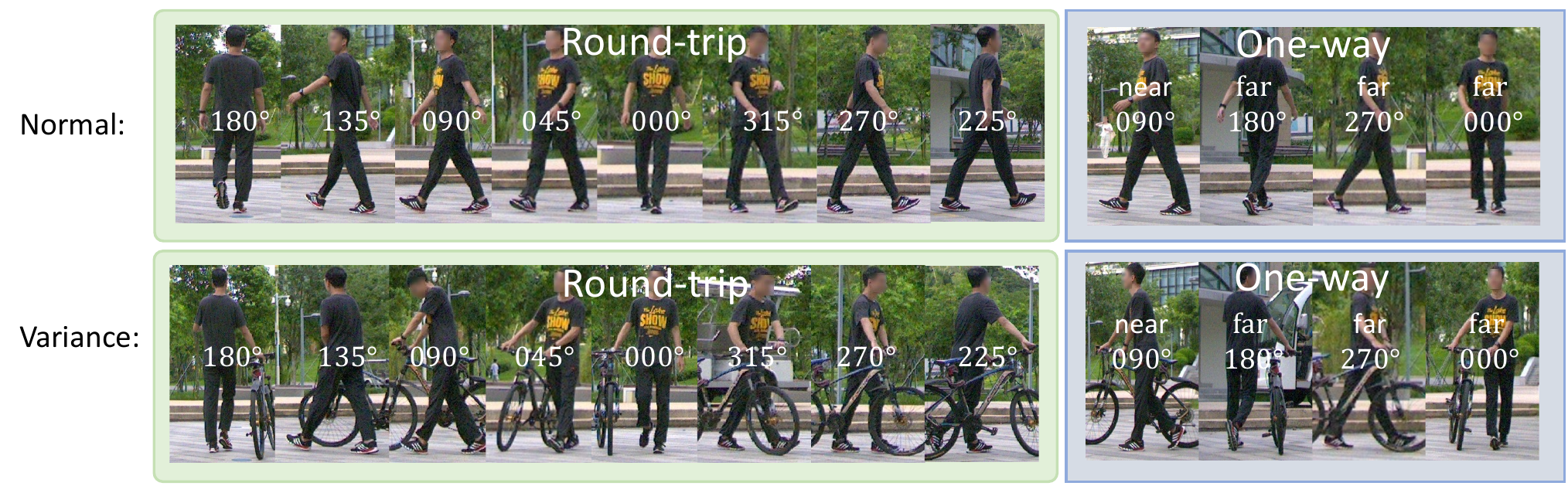}
         \caption{Samples of 12 views in two conditions for a subject.}
         \label{fig:views}
     \end{subfigure}
     \hfill
     \begin{subfigure}[b]{0.25\textwidth}
         \centering
         \includegraphics[width=1\textwidth,height=3.8cm]{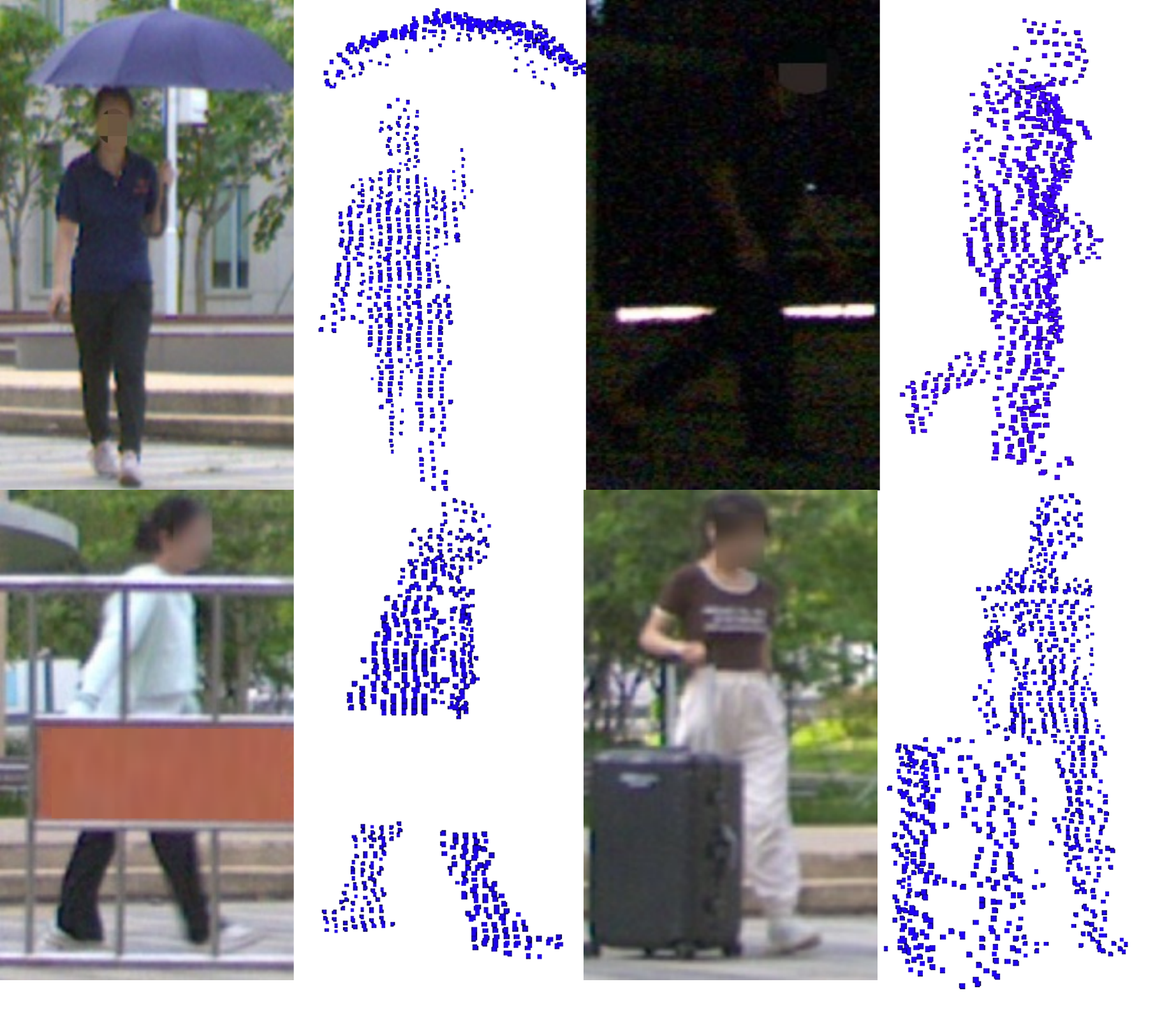}
         \caption{Diverse attributes in SUSTech1K.}
         \label{fig:demo}
     \end{subfigure}
     \hfill

        \caption{(a) Each participant walks normally (top row), followed by walking with a random variance (bike for this subject) as shown in the bottom row. (b) SUSTech1K collects data in point cloud and RGB modality with diverse realistic variances. \label{fig:dataset_preview}}
       \vspace{-4mm}
\end{figure*}

\section{The SUSTech1K Benchmark}

The SUSTech1K dataset is captured by a mobile robot equipped with a 128-beam LiDAR scanner and a monocular camera, providing synchronized multimodal data. It includes 1,050 identities, 25,239 sequences, 763,416 point-cloud frames, and 3,075,575 RGB images with corresponding silhouettes. The SUSTech1K dataset is a synchronized multimodal dataset, with timestamped frames for each modality of frames.
In addition, we protect the privacy of the participants by blurring their faces and obtaining informed consent. 
To the end, we manually annotate various walking conditions in SUSTech1K.

\noindent\textbf{Data Collection.} The dataset was collected in July 2022 in three scenes on the SUSTech campus using an industrial camera and a LiDAR sensor. The camera captured RGB imagery streams at a resolution of $1,280 \times 980$ and 30 frames per second (FPS), while the point cloud streams were recorded at 10 FPS. We synchronized the LiDAR and camera with the GPS clock and timestamped each frame to enable collaboration between the two modalities for a robust gait recognition system. 
The current indoor datasets mainly focus on subject-centered gait recognition with a clean background, while other existing datasets focus on pedestrians from surveillance views. In contrast, SUSTech1K was collected from a robot view.
The existing gait datasets such as OUMVLP~\cite{oumvlp}, and CASIA-E~\cite{casiae} set the capture range to around 8 meters. Our experimental setup is inherited from existing gait datasets at a comparative distance as shown in Fig.~\ref{fig:setup}. Each subject first walked normally along the \textit{one-way paths} and the \textit{round-trip paths}, 
and then walked again with an extra random attribute, such as carrying any object, as shown in Fig.~\ref{fig:views}. Each subject can have a total of 48 gait sequences
\textit{(= [4 $\times$ 2 (round-trip) + 4 $\times$ 1(one-way)] $\times$ 2 (twice) $\times$ 2 (modality))}. 

\noindent\textbf{Variances.} 
% To evaluate the robustness of two modalities in challenging outdoor environments, 
In practice, we instructed each subject to walk with random attributes during their second round. The SUSTech1K dataset preserves the variances found in existing datasets, such as \textit{Normal, Bag, Clothes Changing, Views} and \textit{Object Carrying}, while also considering other common but challenging variances encountered outdoors, including \textit{Occlusion, Illumination, Uniform, and Umbrella}. By categorizing these walking sequences into different subsets based on their variances, we can further explore the impact of different variations on the gait recognition performance of the two modalities.

% \subsubsection{}
\noindent\textbf{Annotations and Representations.} The continuous data streams are first manually segmented into sequences based on the predefined trajectories shown in Fig.~\ref{fig:setup}. Each sequence is then labeled according to the aforementioned variances.
% , such as views, occlusion, and illumination. 
Finally, the camera-based and lidar-based sequences are processed separately to obtain gait representations.

For \textit{camera-based representations,} we first applied human detection~\cite{ge2021yolox}, tracking~\cite{zhang2022bytetrack}, and segmentation~\cite{liu2021paddleseg} on the raw RGB imagery streams, to generate camera-based gait representations with RGB images and corresponding silhouettes. In cases where the tracking algorithm produced inaccurate bounding boxes, we manually corrected them. It should be noted that the performance of the segmentation algorithm deteriorates in low-light conditions, resulting in suboptimal performance. 
% Besides, SUSTech1K was captured on a public campus, increasing the likelihood of passersby in the background and presenting a challenge for foreground segmentation.

For \textit{LiDAR-based representations,} obtaining LiDAR-based representations was relatively easy because there was only one subject walking in the experimental area at a time. To protect the privacy of uninvolved passers and to generate clean gait representations in the point cloud format, we only release the area range to $[-5, -12m]$ for the X axis, $[-3m, 3m]$ for the Y axis, and $[-2m,3m]$ for the Z axis. 
Moreover, 
 % To obtain the final LiDAR-based gait representations, 
we applied noise removal~\cite{dbscan} and ground removal~\cite{groundremoval} on each frame to clean the lidar data.

\begin{figure*}[t]
     \centering
     \begin{subfigure}[b]{0.33\textwidth}
         \centering
         \includegraphics[width=1\textwidth]{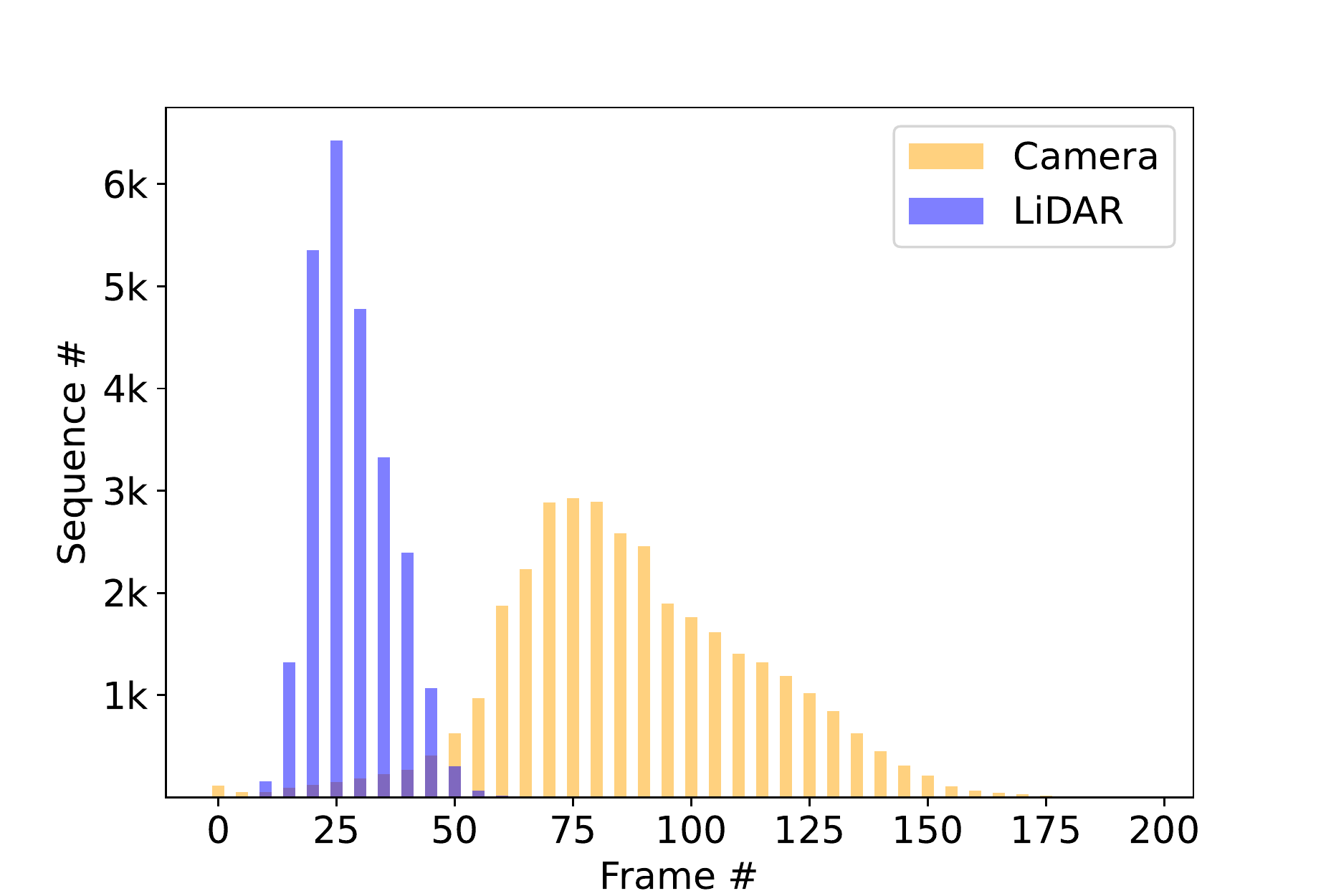}
         \caption{sequence \# over frame \#}
         \label{fig:sequenceframe}
     \end{subfigure}
     \hfill
     \begin{subfigure}[b]{0.33\textwidth}
         \centering
         \includegraphics[width=1\textwidth]{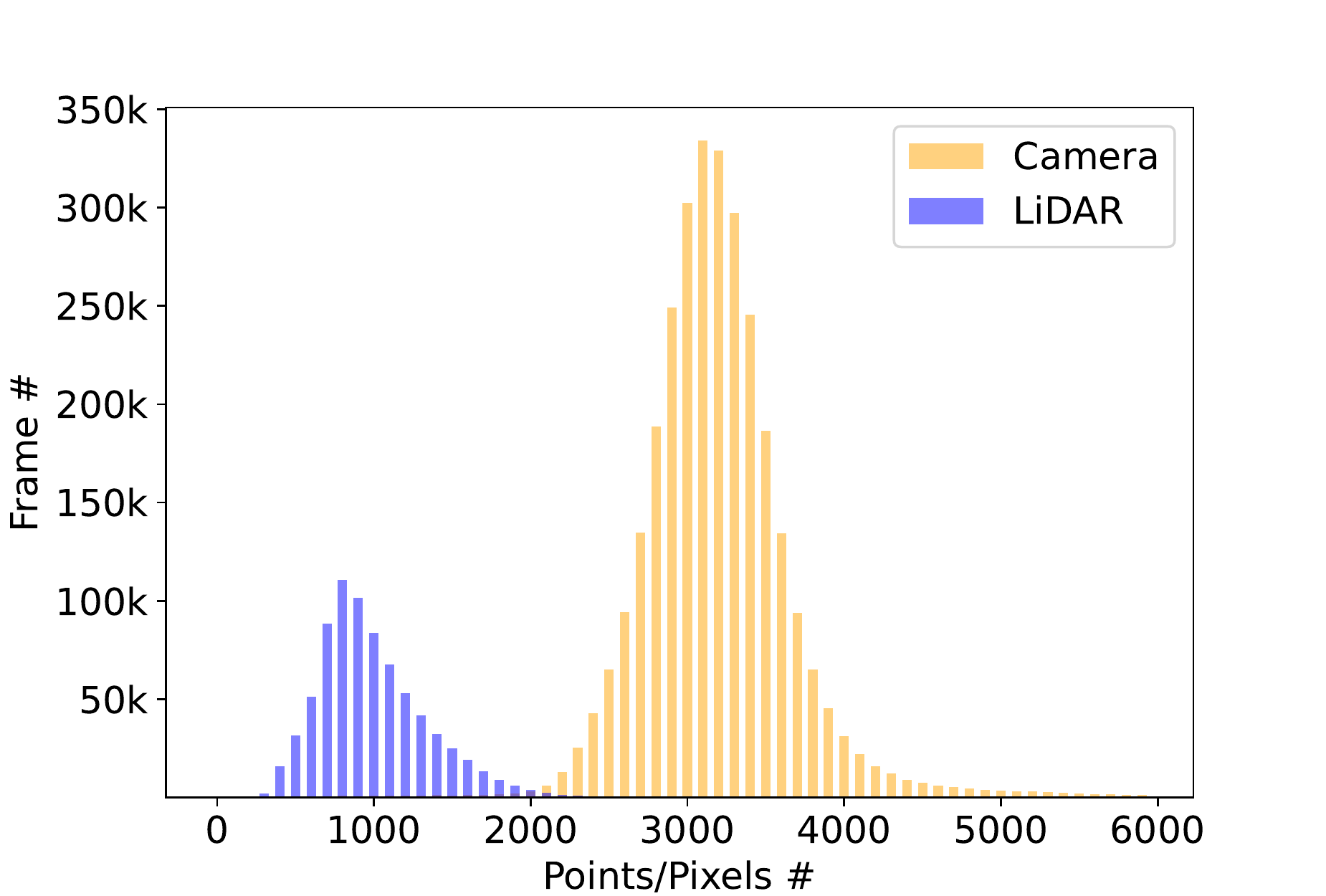}
         \caption{frame \# over points/pixels \#}
         \label{fig:framepoints}
     \end{subfigure}
     \hfill
     \begin{subfigure}[b]{0.33\textwidth}
         \centering
         \includegraphics[width=0.95\textwidth]{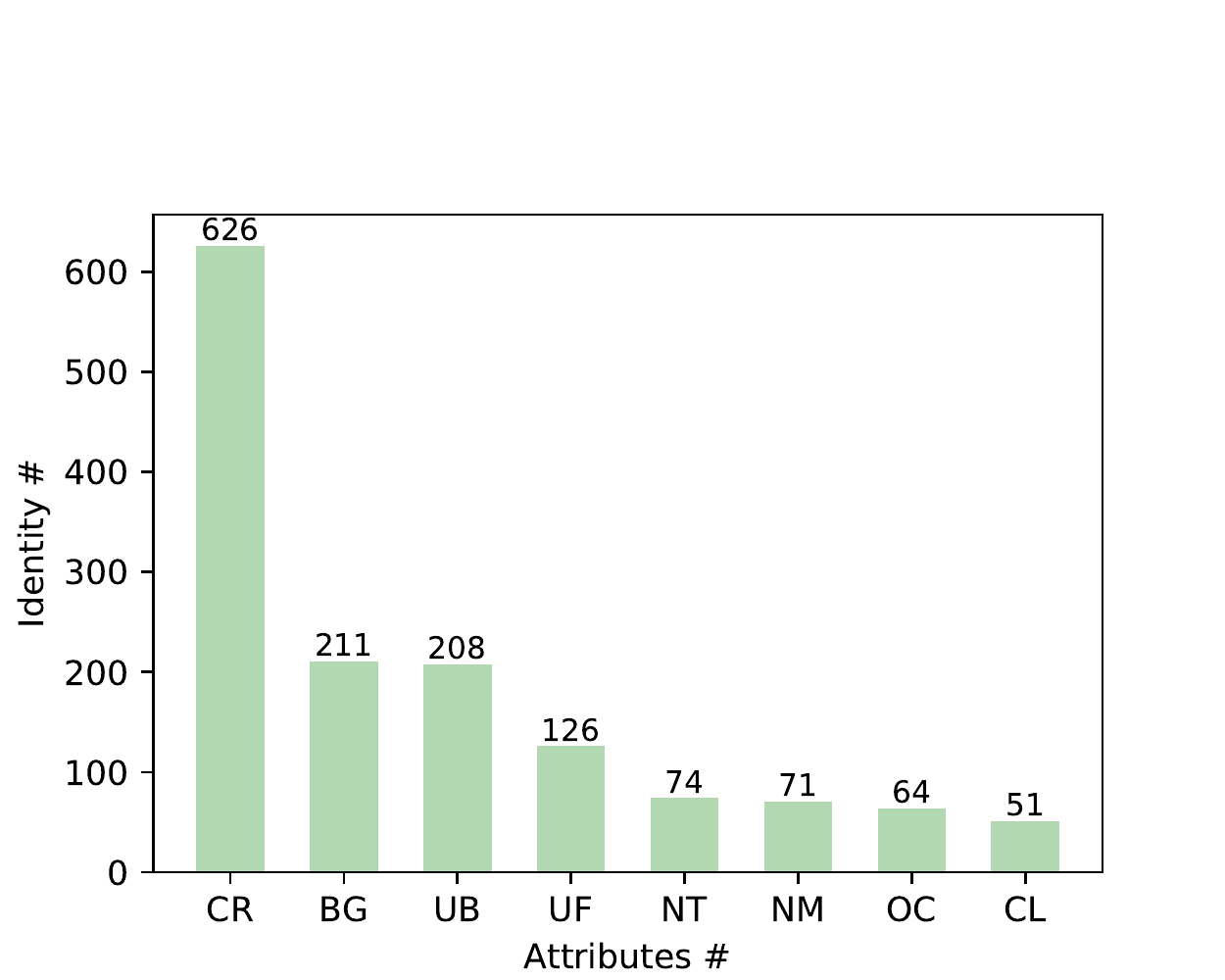}
         \caption{Attributes distribution}
         \label{fig:distribution}
     \end{subfigure}
        \caption{Statistics about SUSTech1K dataset. LiDAR modality and RGB modality are represented in blue and yellow, respectively. It shows that SUSTech1K dataset is scalable, multimodal, and diverse for the study of 3D gait recognition. \textit{CR, BG, UB, UF, NT, NM, OC,} and \textit{CL} denote attributes of Carrying, Bag, Umbrella, Uniform, Night, Occlusion, and Clothing, respectively. Best viewed in color.}
        \label{fig:three graphs}
        \vspace{-3mm}
\end{figure*}

\noindent \textbf{Statistics.}
Fig.~\ref{fig:sequenceframe} indicates that the two modalities have the same number of sequences, while the RGB modality has three times more frames per sequence than the LiDAR modality.
The imagery gait representations provide more details and dense information when the camera-based representations are in the resolution of $128\times128$ as shown in Fig.~\ref{fig:framepoints}. When we resize the imagery to the resolution of $64\times64$, the ratio of pixels vs points is approximately 1:1, allowing for a more direct and fair comparison of the two modalities.
In the end, the distribution of attributes in Fig.~\ref{fig:distribution}, shows the diversity of the SUSTech1K dataset.

\noindent \textbf{Evaluation Metrics.} To establish a more challenging and realistic setting, the SUSTech1K dataset is evaluated under an open-set setting~\cite{nixon2010book,oumvlp}, where train and test set splits are without sample overlapping. The evaluation protocol follows the cross-view recognition setting as commonly used in CASIA-B~\cite{casiab} and OUMVLP~\cite{oumvlp}, where probe sets of the same view calculate the similarity to gallery sets of each view. The probe sets are grouped into many subsets according to the attributes to evaluate the impact of attributes, then perform cross-view retrieval task. The prevailing Rank-1 accuracy and Rank-5 are adopted as the evaluation metric.

\section{Gait Recognition with Point Clouds}

\subsection{Problem Setting}
In this section, we introduce the problem setting of 3D gait recognition with point clouds. Given a point cloud dataset $\mathcal{P}=\{\mathcal{P}_i^j|i=1,2...,N; j=1,2,...,m_i\}$ with N identities and $m_i$ sequence for each identity $y_i$. Each point cloud sequence $\mathcal{P}_{i}^{j} \in \mathbb{R}^{T \times N \times C} $ is with $T$ frames and $N$ points for each frame, where $C$ is the number of feature channels. Our goal is to learn a network $N_\theta(\cdot)$ that can produce the feature embedding $\mathcal{F}_{i}^{j}$ to represent the associated identity $y_i$. 

\begin{figure*}[ht!]
\centering
\includegraphics[width=0.8\linewidth]{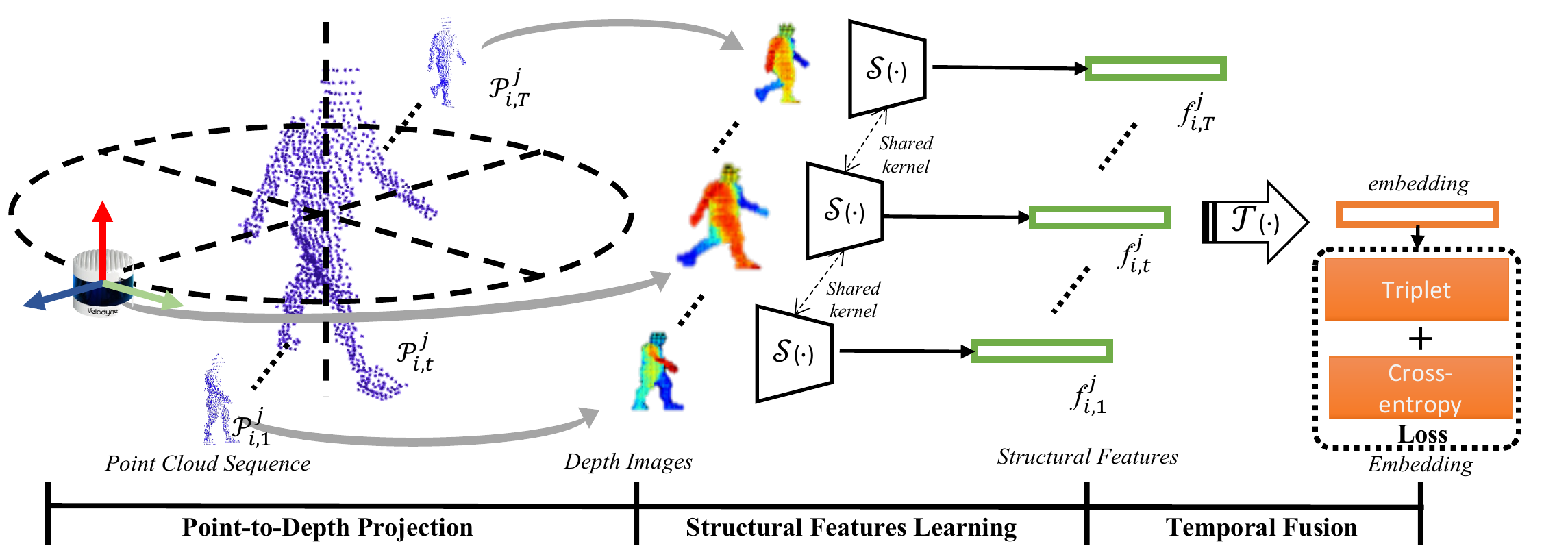}
\caption{The framework of LidarGait for 3D gait recognition with point clouds. LidarGait receives a sequence of point clouds, extracts representations from range-view projection depths, and aggregates sequential features by set pooling.  }
\label{fig:method}
\vspace{-4mm}
\end{figure*}

We propose the LidarGait, as shown in Fig.~\ref{fig:method}, to tackle the 3D gait recognition task, formulated as:
\begin{equation}
    \mathcal{F}_{i}^{j} = N_\theta(\mathcal{G}(\mathcal{P}_i^j))
    \vspace{-1mm}
\end{equation}
where the projection function $\mathcal{G}$ operates on point clouds and generates depth images from the LiDAR front view.
 The feature extractor $N_\theta$ is composed of two components. 1) a structural feature encoder $\mathcal{S}$ that captures spatially local connectivity from projected front-view depth images. 2) a temporal aggregation network $\mathcal{T}$ that models dynamical conjunction along sequential input, which can be formulated as:
\begin{equation}
    N_\theta(\cdot) = \mathcal{T}(\mathcal{S}(\mathcal{G}(\mathcal{P}_i^1), \cdots, \mathcal{S}(\mathcal{G}(\mathcal{P}_i^{m_i})))
    \vspace{-1mm}
\end{equation}

LidarGait first receives sequential 3D point clouds and then extracts spatial-temporal representation from projected depth maps. To end, LidarGait is optimized by combining triplet and cross-entropy loss.

\subsection{LidarGait}
\label{sec:lidargait}

\noindent\textbf{Point-to-Depth Projection.} The range-scanned point clouds from a Velodyne VLS128 LiDAR scanner, can be projected and discretized into a 2D point map, using the following projection function~\cite{point2depth}:
\begin{equation}
\begin{aligned}
    r &= \lfloor \textrm{atan2} (y, x) / \Delta \theta \rfloor \\
    c &= \lfloor \arcsin (z / \sqrt{x^2 + y^2 + z^2}) / \Delta \phi \rfloor
\end{aligned}
\label{eq:projection}
\end{equation}
where 3D point $\mathbf{p} = (x, y, z)^\top$ is mapped to its corresponding 2D pixel coordinates $(r, c)$ in the projected depths image. The $\Delta \theta$ and $\Delta \phi$ represent the average horizontal and vertical angle resolution between consecutive beam emitters.  According to the configuration of the LiDAR, the $\Delta \theta$ and $\Delta \phi$ are set to 0.192 and 0.2, respectively. The resulting 2D point map is similar to cylindrical images. Each element in the map at position $(r, c)$ is filled with $d$, where $d = \sqrt{x^2 + y^2}$. In the rare case where multiple points are projected to the same 2D position, only the point closest to the observer is kept. If no 3D point is projected onto a particular 2D position, the corresponding element in the point map is filled with $0$. Then the depth projection is normalized and converted to RGB images from the 1-channel images.

\noindent\textbf{Structural Representation Learning.} LidarGait extracts abstract structural features from sequences of depth images using a convolutional network $\mathcal{S}$. The $\mathcal{S}$ is a spatial feature encoder that can use any existing silhouette-based backbone. In this work, we use GaitBase~\cite{opengait} as our feature encoder $\mathcal{S}$.
As opposed to camera-based methods that use silhouettes as input, point-wise methods~\cite{pointnet,pointnet++,pointtransformer} extract representation directly from point clouds. However, point-wise methods underperform camera-based recognition methods as shown in Tab.~\ref{tab:attributes}, despite using the informative 3D structures of pedestrians. We attribute this performance gap to the fact that current point-based models are optimized for global feature modeling, which is suitable for distinguishing objects with large inter-class differences. However, gait recognition requires capturing fine-grained features to distinguish individuals with small inter-class distances and large intra-class distances. To address this challenge, LidarGait utilizes convolutional networks to extract gait representations from projection depths, which are better at capturing such fine-grained features.

\noindent\textbf{Temporal Fusion.} To aggregate the features from the variable length of depth sequences, we use Set Pooling~\cite{gaitset} as the temporal feature aggregator, which enables the model to capture the final sequence-level gait representation.

\begin{figure}[t]
\centering
\includegraphics[width=0.75\linewidth]{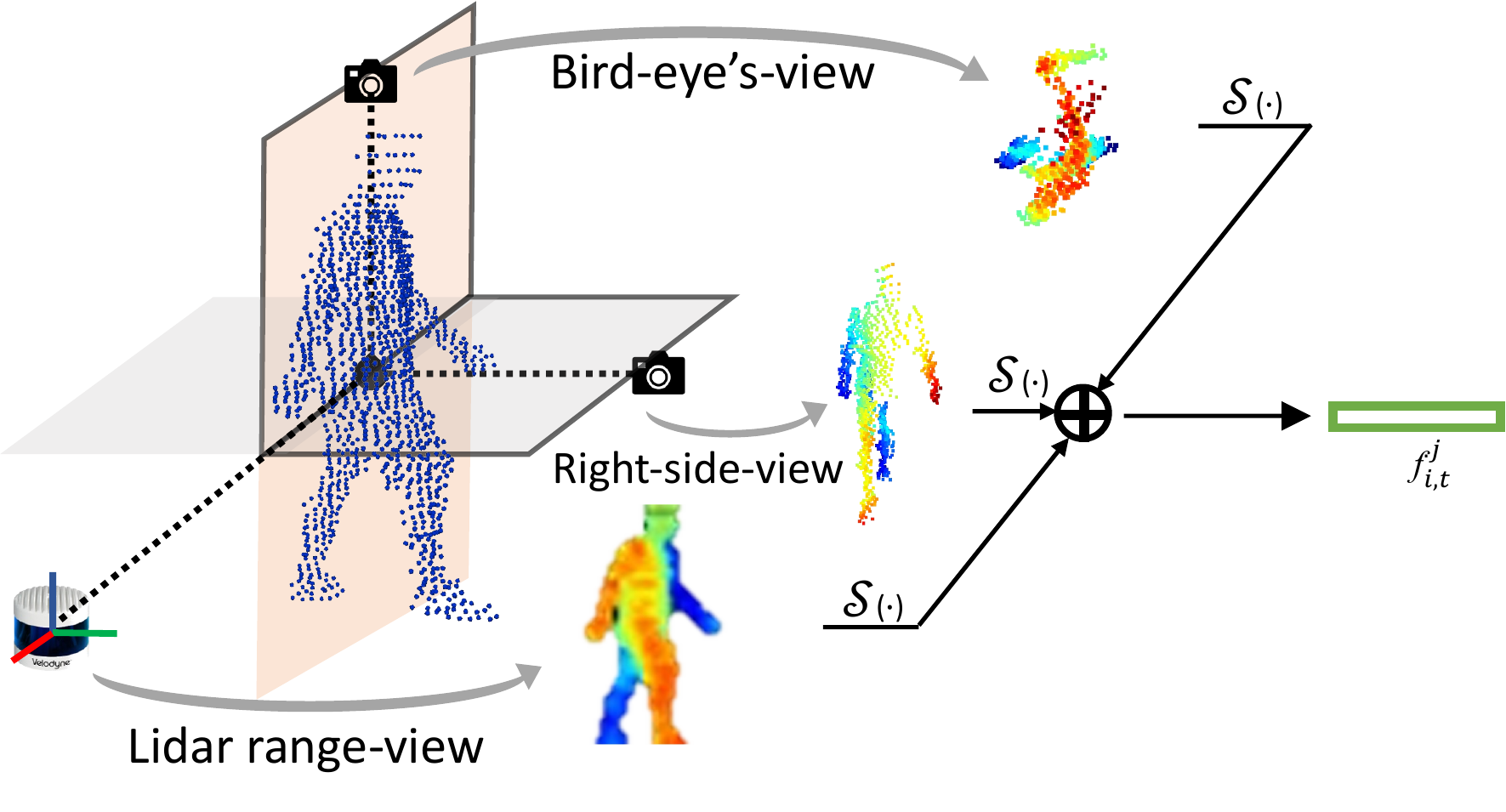}
\caption{Illustraition of MV-LidarGait, which aggregates extra depths projection from the right-side and bird-eye view.}
\label{fig:mvlidargait}
\vspace{-5mm}
\end{figure}

\noindent\textbf{MV-LidarGait.} In addition to obtaining projected depths from the perspective of the LiDAR sensor, point clouds can also be projected from orthographic views as described in SimpleView~\cite{simpleview}. To verify whether the LiDAR range-view projection is adequate and to explore the effectiveness of other projected views, we extend LidarGait to MV-LidarGait, which projects point sets into two extra orthogonal views, as illustrated in Fig.~\ref{fig:mvlidargait}.  Each rendered view is independently extracted by a feature encoder and fused in a frame-by-frame manner. 

\subsection{Traning and Inference}
The model is trained using a combined loss function that includes the $BA^{+}$ triplet loss~\cite{gaitset} and the cross-entropy loss~\cite{gaitgl}, with weighted hyperparameters $\alpha$ and $\beta$, respectively:
\begin{equation}
L = \alpha L_{tri} + \beta L_{ce}
\end{equation}

During inference, the similarity between each probe-gallery pair is measured using the Euclidean distance. %, and the Rank 1 recognition accuracy is calculated.

\section{Experiments}

\subsection{Experimental Setup}
\noindent\textbf{Evaluation Protocol.} All experiments are conducted on the SUSTech1K dataset, which is divided into three splits: a training set with 250 identities and 6,011 sequences, a validation set with 6,025 sequences from 250 unseen identities, and a test set with the remaining 550 identities and 13,203 sequences. The SUSTech1K dataset provides gait sequences from multiple viewpoints, enabling the study of cross-view gait recognition in both camera and LiDAR modalities. The cross-view evaluation protocol~\cite{casiab,oumvlp} in CASIA-B and OUMVLP is adopted for SUSTech1K as well. During the test, the sequences in normal conditions are grouped into gallery sets, and the sequences in variant conditions are taken as probe sets. 

\noindent\textbf{Evaluation on Each Condition.} To investigate the impact of various realistic factors on gait recognition in the wild, including clothes changing, poor illumination, object carrying, occlusion, and wearing uniforms, we group all probes with the different covariates into multiple subsets for evaluation. For instance, the umbrella subset consists of probes with an umbrella to evaluate the effect of carrying an umbrella, with the gallery set containing all sequences in normal conditions.

\subsubsection{Comparative Methods}
As detailed in Sec.~\ref{sec:lidargait}, LidarGait utilizes the GaitBase with set pooling to capture 3D gait features on range-view depths. We evaluate LidarGait with the below methods.

\noindent\textbf{Camera-based Methods.}
To evaluate the performance of the camera-based modality, we implement four cutting-edge methods: GaitSet~\cite{gaitset}, GaitBase~\cite{opengait}, GaitPart~\cite{gaitpart}, and GaitGL~\cite{gaitgl}. The network parametric setting is identical to the configuration for CASIA-B, which has the equivalent scale of the training set to SUSTech1K.

\noindent\textbf{Lidar-based Methods.}
We implement four commonly used approaches in point cloud classification including PointNet~\cite{pointnet}, PointNet++~\cite{pointnet++}, PointTransformer~\cite{pointtransformer}, and SimpleView~\cite{simpleview}.
Among them, the first three methods~\cite{pointnet,pointnet++,pointtransformer} are point-wise models, while SimpleView~\cite{simpleview} is a representative projection-based method.

\begin{table*}[t]
\centering
\caption{Evaluation with different attributes on SUSTech1K \textit{valid + test} set. The bolded and underlined values represent the first and second-best results, respectively.
\label{tab:attributes}}
\vspace{-3mm}
    \scalebox{0.75}{%

        \begin{tabular}{lcc|cccccccc|c|c}
        
        \midrule[1.5pt]
        \multirow{2}{*}{Model}   & \multirow{2}{*}{Publication}    &  \multirow{2}{*}{Modality}     & \multicolumn{8}{c}{Probe Sequence (\textit{Rank-1} \textit{acc})}      &  \multicolumn{2}{|c}{Overall}                    \\
          &                       &                                              & Normal    & Bag    & Clothing  & Carrying  & Umbrella  & Uniform  & Occlusion & Night   & \textit{Rank1}    & \textit{Rank5}  \\         \midrule[1pt]
        GaitSet~\cite{gaitset}      & AAAI2019    & \multirow{4}{*}{Camera}       & 69.10     & 68.25   & 37.44     & 65.01     & 63.08      & 61.00    & 67.19     & 23.04   & 65.04  & 84.76  \\
        GaitPart~\cite{gaitpart}    & CVPR2019    &                             & 62.20     & 62.81   & 33.08     & 59.53     & 57.25      & 54.85    & 57.20     & 21.75   & 59.19  & 80.79  \\
        GaitGL~\cite{gaitgl}        & ICCV2021    &                             & 67.11     & 66.16   & 35.92     & 63.31     & 61.58      & 58.07    & 66.59     & 17.88  &  63.14 & 82.82  \\ 
        GaitBase~\cite{opengait}    & CVPR2023    &                             & \underline{81.46}     & \underline{77.48}   & 49.60     & \underline{75.77}     & \textbf{75.55}      & \underline{76.66}    & \underline{81.40}     & 25.92   & \underline{76.12}  & \underline{89.39}   \\
                                     \hline 
        PointNet~\cite{pointnet}    & CVPR2017    &  \multirow{4}{*}{LiDAR}     & 43.59     & 37.27   & 25.72    & 28.78     & 19.85       & 30.05    & 44.29     & 27.35   & 31.33  & 59.75  \\
        PointNet++~\cite{pointnet++}& NIPS2017    &                             & 55.90     & 52.22   & 41.60    & 49.60     & 47.84       & 45.91    & 54.16     & 52.49   & 50.78  & 82.38 \\
        PointTransformer~\cite{pointtransformer}& ICCV2021     &                & 53.19     & 48.08   & 32.05    & 43.20     & 39.06       & 41.75    &  47.87    & 47.12   &  44.37  & 76.70 \\ 
        SimpleView~\cite{simpleview}& ICML2021    &                             & 72.33     & 68.75   & \underline{57.15}    & 63.26     & 49.20       & 62.52    & 79.72     & \underline{66.54}   & 64.83  & 85.77   \\  \hline 
        \textbf{LidarGait}         & Ours        &      LiDAR                  & \textbf{91.80}     & \textbf{88.64}   & \textbf{74.56}    & \textbf{89.03}     & \underline{67.50}       & \textbf{80.86}    & \textbf{94.53}     & \textbf{90.41}   & \textbf{86.77}  & \textbf{96.08} \\ 
        \midrule[1.5pt]
        \end{tabular}
        }

\vspace{-4mm}
\end{table*}

\noindent\textbf{Implementation Details.} All the camera-based silhouettes and LiDAR-based depth images are aligned using the method introduced in~\cite{oulp} and then resized in the resolution of $64\times64$. For LiDAR modality methods, we use the SGD optimizer with a weight decay of 0.0005 and an initial learning rate of 0.1. The learning rate is reduced by a factor of 0.1 at the 20,000th and 30,000th iterations, and the total number of iterations is set to 40,000. For methods using camera modality, the Adam optimizer is used to prevent the issue of gradient vanishing because of low-quality silhouettes.  The triplet and cross-entropy loss weights are set to 1 and 0.1, respectively. The batch size ($p$, $k$, $l$) is set to ($8$, $8$, $10$) and ($8$, $16$, $30$) for lidar-based methods and camera-based methods, respectively, where $p$ denotes the number of IDs, $k$ for the number of sequence of training samples per ID, and $l$ is the number of frame per sequence. All comparison methods are trained using the same training strategy as LidarGait. The OpenGait~\cite{opengait} codebase is used to conduct all experiments\footnote{\url{https://github.com/ShiqiYu/OpenGait}}.

\noindent\subsection{Comparative Results}
\begin{figure}[t]
\centering
\includegraphics[width=1\linewidth]{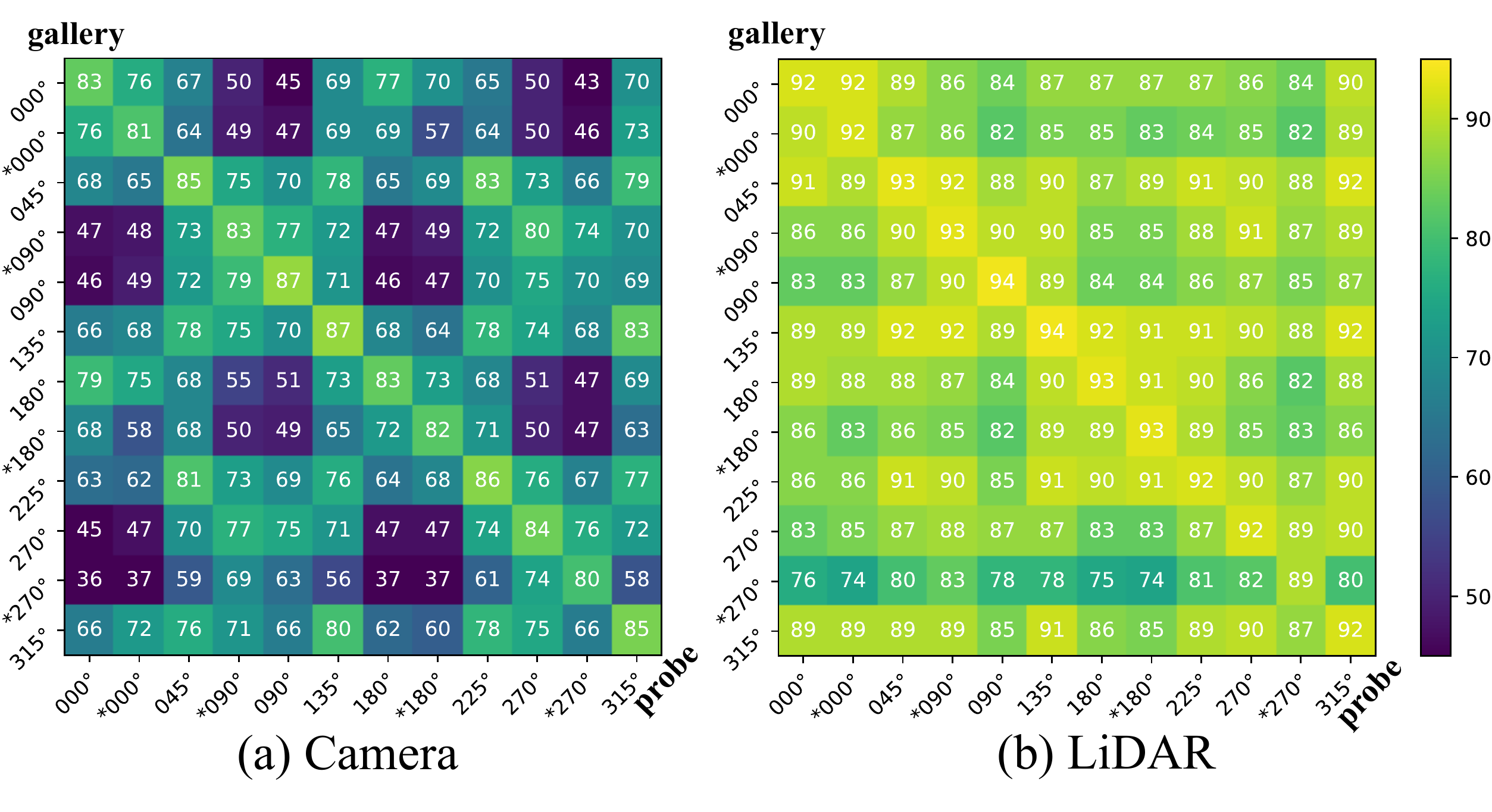}
\vspace{-8mm}
\caption{Cross-view performance comparison between LiDAR and camera for gait recognition. We report rank-1 accuracy (\%) on the cross-view protocol. * indicates viewpoint at a longer distance.  Best viewed in color and pdf.}
\label{fig:comparison}
\vspace{-4mm}
\end{figure}

 Following the cross-view evaluation protocol~\cite{casiab,oumvlp}, we evaluate all methods on each subset with different conditions, and we report the cross-view accuracy matrix in Fig.~\ref{fig:comparison} for a detailed performance comparison between LiDAR and camera modality. We report the average of the accuracy matrix in Tab.~\ref{tab:attributes}, obtaining the following observations: (1) LidarGait shows its superiority to all existing point-based and camera-based methods, which is mainly beneficial by integration with 3D geometry information. (2) LidarGait achieves state-of-the-art results in all conditions except the umbrella subset. It is mainly caused that umbrellas are erased after segmentation on RGB images, while the umbrellas are kept in point sets. (3) The methods using silhouettes make a poor performance at night. Point-based methods provide more promising results, and LidarGait outperforms others by a large margin. (4) All silhouette-based models~\cite{gaitset,gaitpart,gaitgl,opengait} achieve higher accuracy than point-based models~\cite{pointnet,pointnet++,pointtransformer} in point cloud classification. This concludes that it is necessary to design point-based models for 3D gait recognition specifically. 
 (5) The models utilizing the order of the frames in sequences, \ie GaitPart, and GaitGL, obtain lower results. While other set-based methods, \ie GaitSet, and GaitBase, perform better accuracy. It means that temporal cues may be impacted in the outdoor scenes because of low-quality silhouettes.

\noindent\textbf{Cross-view Gait Recognition.} We conduct a detailed comparison of two modalities of cross-view gait recognition in Fig.~\ref{fig:comparison}. The identical feature encoder, GaitBase, is utilized for two modalities to make ablative results. We can make the following observations: (1) the distance from subjects to sensors indeed impacts the performance for both two modalities. (2) Camera-based method achieves poor performance when query sets are at views of $0^{\circ}, 90^{\circ}, 180^{\circ}$ (see purple pixel in Fig.~\ref{fig:comparison}\textcolor{red}{a}). The same phenomenon can be found on CASIA-B~\cite{gaitset} and OUMVLP~\cite{oumvlp}. However, LiDAR-based methods can perform stably in cross-view settings, validating the effectiveness of 3D structure for cross-view gait recognition.

\subsection{Ablation Study}

\begin{figure}[t]
\centering
\includegraphics[width=0.85\linewidth]{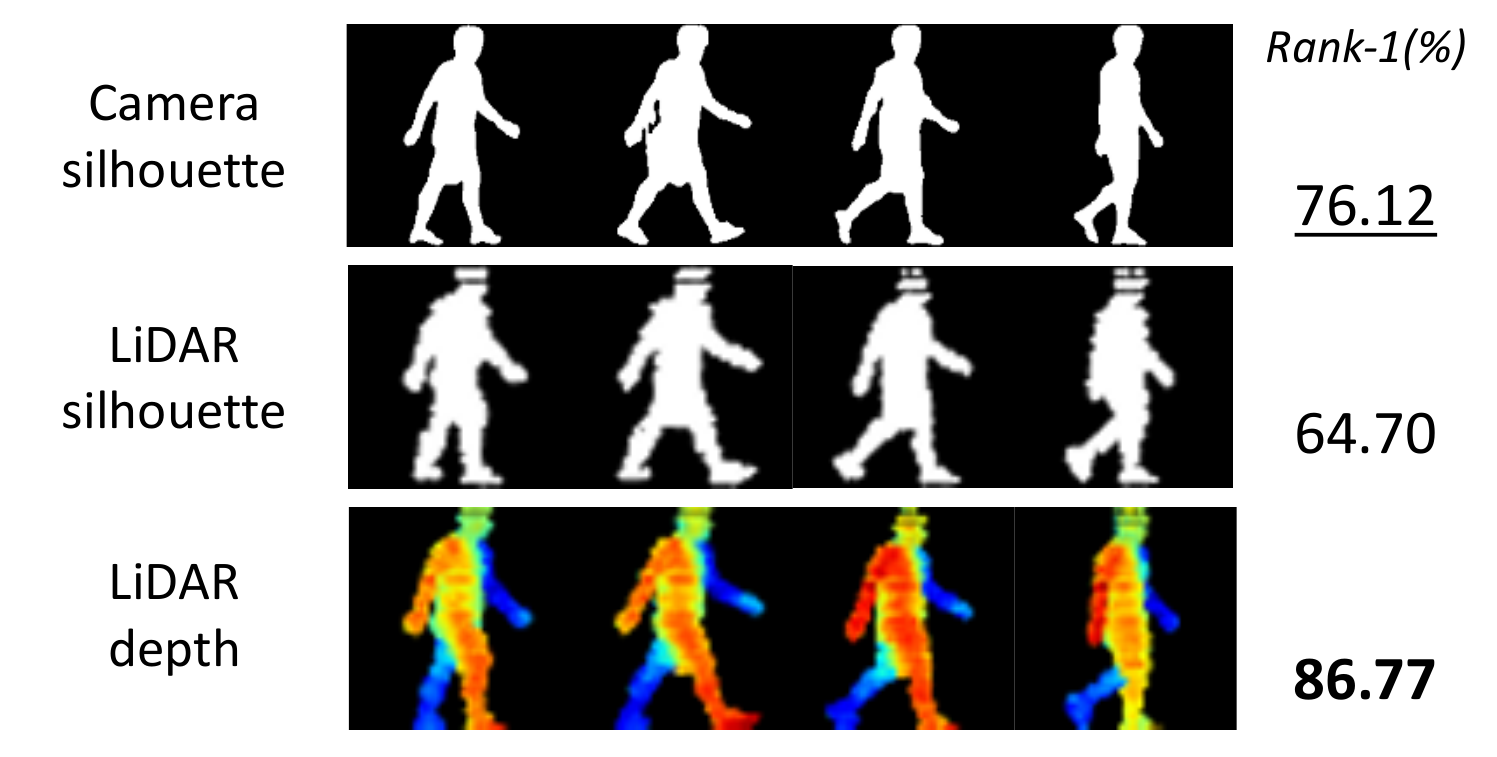}
\vspace{-3mm}
\caption{Ablation study on the effectiveness of depth information for performance. Best viewed in color.\label{fig:input}}
\vspace{-4mm}
\end{figure}

\noindent\textbf{Effectiveness of 3D Geometry Information.}
To evaluate the effectiveness of depth information for gait recognition, we compare four types of data as input: (1) Camera silhouettes: the camera-based silhouettes are obtained by segmentation results of RGB images. (2) LiDAR silhouettes: LiDAR silhouettes are obtained by range-view projection of point cloud sets.  (3) LiDAR depth: the depth information is added.  From Fig.~\ref{fig:input}, we can observe that: (1) When depth information is not included, the performance of LiDAR silhouettes is much lower than the accuracy of camera silhouettes. This is because the camera has a much higher resolution than Lidar, so the silhouettes from the camera can have more details. (2) Though LiDAR generates point clouds in sparse space, the depth of information makes a magnificent improvement to the accuracy. Integrating depth information can improve rank-1 accuracy from 64.70\% to 86.77\%, validating the necessity and effectiveness of 3D information for gait recognition.

% \noindent\textbf{The Impact of Frames length.}

\noindent\textbf{Effectiveness of Other Projected View.}
% To study the contributions of different views on the performance of 3D gait recognition, we train with each view separately and then evaluate their performance of gait recognition. 
The right-side view (\textbf{RSV}) is obtained by positioning a virtual camera orthogonal to the LiDAR range view (\textbf{RV}) on the right-hand side. The Bird-eye's view (\textbf{BEV}) projects point clouds onto a plane above the point clouds of pedestrians. Based on the results presented in Tab.~\ref{tab:mvlidargait}, we have the following observations: 
(1) LidarGait achieves the best performance on the range view projection when a single viewpoint is used as input.
(2) The right-side view also provides a reliable representation, which performs comparably to camera-based methods.
(3) Although the accuracy of Bird's-eye views is only 26.33\%, learning gait features from BEV images provides interesting evidence that gait recognition can potentially be achieved at the Bird's-eye view.
(4) MV-LidarGait can be improved (\textit{+0.73\%}) from LidarGait by combining multiple viewpoints, with the improvement mainly coming from the umbrella subset.
(5) The integration of BEV does not enhance the performance of MV-LidarGait, indicating that BEV only provides redundant information already contained in the other two viewpoints.  

More experiments and exemplar data on SUSTech1K are included in \textbf{the supplementary material}\footnote{\url{https://lidargait.github.io/}}.

\section{Discussion}

\noindent\textbf{Ethical Discussion.} The SUSTech1K dataset has been reviewed by the Southern University of Science and Technology Institutional Review Board (SUSTech IRB). All the subjects involved in the dataset signed a written consent to agree that their data can be collected, processed, used, and shared for research purposes. The dataset can be distributed only for non-commercial research purposes with the case-by-case dataset access application. The human faces are blurred from RGB images to protect sensitive privacy. The recorded data can only be used for 20 years since this paper publishes.
After this date, all data will be deleted and not allowed to be used.

\begin{table}[t]
\centering
\caption{Effectiveness of each projected view. MV-LidarGait achieves the best performance.
\label{tab:mvlidargait}}
\vspace{-3mm}
\scalebox{0.72}{
\begin{tabular}{ll|rr|r}
\midrule[1.5pt]
\multirow{2}{*}{Model}                      & \multirow{2}{*}{Used views} & \multirow{2}{*}{Normal} & \multirow{2}{*}{Umbrella} & Overall  \\
                           &      &        &          & \textit{Rank1}        \\\midrule[1pt]
\multirow{3}{*}{LidarGait} & BEV   & 39.92  & 12.12    & 26.33      \\
                           & RSV   & 70.61  & 47.02    & 62.67      \\
                           & RV    & \textbf{91.80}  & 67.50    & 86.77      \\\midrule[1pt]
\multirow{2}{*}{MV-LidarGait}&  RV + RSV  & \underline{91.29}  & \textbf{70.91}    & \textbf{87.50}     \\
                                & RV + RSV + BEV   & 91.22  & \underline{69.43}    & \underline{87.47}    \\
\midrule[1.5pt]
\end{tabular}
}
\vspace{-6mm}
\end{table}

\section{Conclusions and Future work}

In this paper, we introduce the LiDAR sensor to provide reliable anthropometric parameters for the human body, and to perceive pedestrians in unconstrained scenes. First, we proposed a novel multi-view projection network for point cloud gait recognition, named LidarGait, to exploit 3D human geometry from multi-view representations. Moreover, we build the first large-scale multimodal 3D point cloud gait recognition dataset, termed SUSTech1K, to facilitate the research of gait recognition with point cloud data. SUSTech1K contains 25,239 sequences with 1,050 subjects and covers various visibility, views, occlusions, clothing, carry, and scenes. Lastly, our proposed method achieves remarkable results on the SUSTech1K dataset, showing the superiority of LiDAR and the effectiveness of LidarGait. 
    
LidarGait has obtained remarkable results in various scenarios, yet it does not perform well when subjects carry umbrellas. The reason should be that the umbrellas are wrongly included in the point cloud. Better performance can be achieved if the umbrellas are removed from the point clouds. Besides, LidarGait only takes one modality as input currently. SUSTech1K dataset is a multimodal dataset with synchronized RGB images and point clouds. Much better results should be achieved if the two modalities are fused. 

% \noindent \textbf{Acknowledgement.}
\section*{Acknowledgements}
We thank Jingzhe Ma and Junhao Liang for their invaluable assistance in the SUSTech1K collection, and Xinlai Liu for his careful proofreading. We also thank the BDStar company for providing the LiDAR sensor that helps us explore this research. 
This work was supported in part by the National Natural Science Foundation of China under Grant 61976144, 
in part by the Stable Support Plan Program of Shenzhen Natural Science Fund under Grant 20200925155017002, 
in part by the National Key Research and Development Program of China under Grant 2020AAA0140002, 
in part by the Shenzhen Technology Plan Program under Grant JSGG20201103091002008,
and in part by the HKSAR RGC Theme-based Research Scheme under Grant T32-707-22-N.
\newpage

\setcounter{section}{0}
\renewcommand{\thesection}{\Alph{section}}

\section{The Suboptimal Performance on Umbrella Subset}
\begin{figure*}[t]
\centering
\includegraphics[width=0.9\linewidth]{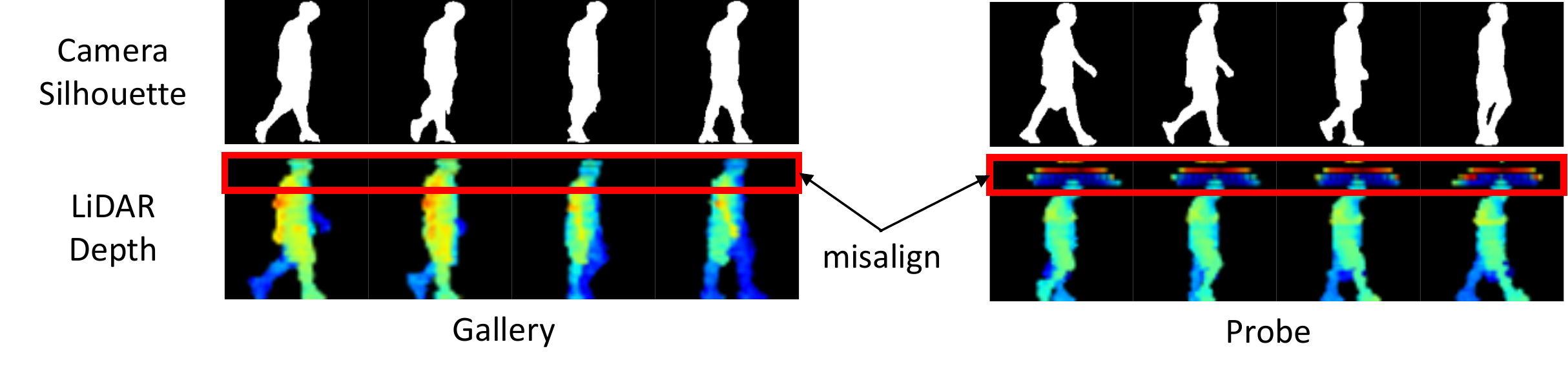}
\vspace{-5mm}
\caption{The exemplar depths and silhouettes from LiDAR and camera modality, where LiDAR modality exists misalignment issue when the probe carries an umbrella.\label{fig:misalgin}}
\vspace{-3mm}
\end{figure*}

Tab. 2 shows that LidarGait outperforms all other methods in all subsets except for cases where pedestrians carry an umbrella. This suboptimal performance is mainly due to the inclusion of the umbrella in the projection images as illustrated in Fig.~\ref{fig:misalgin}, which causes misalignment issues decreasing performance. To improve performance in the umbrella subset, we suggest exploring approaches such as umbrella removal or training with random erasing.

\section{Multi-view Fusion}
In this work, we proposed LidarGait and an extended MV-LidarGait. MV-LidarGait incorporates point clouds from multiple perspectives to generate various depths and fuse multiple depths from different viewpoints into compact multi-view features. As experiments conducted in Tab. 3, we observed that BEV did not improve the performance, and thus, we solely focused on feature fusion for the front-range and right-side views. We explored two methods for combining multiple features: concatenation and sum, and also investigated sequence-level and frame-level feature fusion.  Specifically, multi-view features could be fused in a frame-by-frame manner or after the temporal fusion of each viewpoint branch. The results revealed that only frame-level concatenation led to improved performance, which we report in Tab. 3.

\section{Effect of Sequence Length}

We have studied the impact of sequence length on the inference process. During the inference stage, we examine different frame numbers as input. The frames are randomly selected from the sequence instead of continuous frames sampling. We can observe that: (1) Both camera-based and Lidar-based models obtain better performance with the increasing frames of the sequences. (2) When only given one frame for each sequence of probe and gallery, the LiDAR-based method can surprisingly achieve $25.82$ \% rank-1 recognition accuracy and $52.29$ \% rank-$5$ recognition accuracy, showing the effectiveness of Lidar-based gait recognition in the few-shot setting. (3) The rank-1 accuracy made by LiDAR can be compared to the rank-5 result using camera modality, demonstrating the superior performance of LiDAR in the content of gait recognition. 

	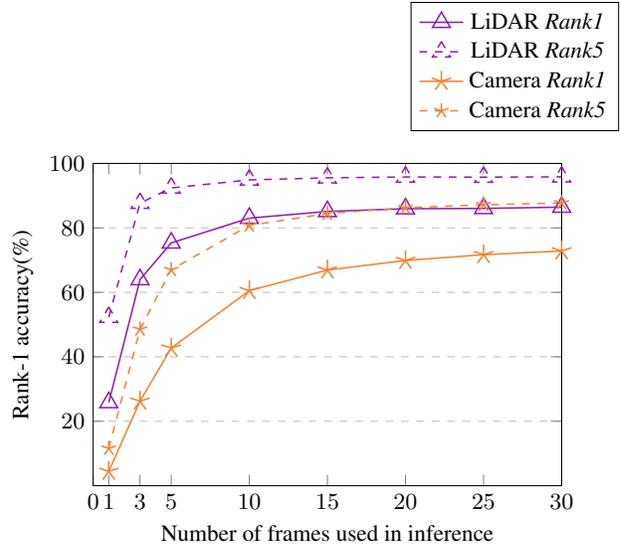
\begin{figure}[ht] %插入图片
		\centering %图片居中
		\resizebox{1\columnwidth}{!}{  %用于修改图片大小
			\begin{tikzpicture} %tikz图片
			\scalefont{0.9} %设置字体大小
			\begin{axis}[
			sharp plot, %控制线的风格
			% title=line chart,%图像标题
			xmode=normal,% 控制坐标轴为线性
%		ymode=log,% 控制坐标轴为对数
			xlabel={Number of frames used in inference}, %x坐标名
			ylabel={Rank-1 accuracy(\%)}, %y坐标名
			width=8cm, height=6cm,  %设置长和宽
			xmin=0,xmax=30,  % 设置x坐标范围
			ymin=0, ymax=100,  % 设置y坐标范围
			xtick={0,1,3,5,10,15,20,25,30}, %指定x轴刻度值。如果为空，则自动设置刻度线。即分割坐标轴
			ytick={20,40,60,80,100}, %指定y轴刻度值。如果为空，则自动设置刻度线。即分割坐标轴
			xlabel near ticks, % 设置x坐标名位置靠近折线图
			ylabel near ticks, % 设置y坐标名位置靠近折线图
			ymajorgrids=true, % 启用/禁用 [公式] 轴上刻度线位置上的网格线
			grid style=dashed, % 设置网格线格式
			legend style={at={(0.9,1.1)},anchor=south}, % 设置标签位置
%			legend columns=3, %设置标签列数
%			legend pos=north west, % 设置折线对应标签的位置
%			legend style={nodes={scale=0.6, transform shape}},  % 设置折线标签的格式
			]
			
			%画第一条线，semithick设置线的粗细为0.6pt，mark是折线标示形状，options是mark形状的大小 ， olive!50!white是颜色，coordinates中包含要绘制的点的坐标
			\addplot+[semithick,mark=triangle,mark options={scale=2}, color=color1] plot coordinates { 
				(1,25.82)
                    (3,63.96)
				(5,75.34)
				(10,83.03)
				(15,85.10)
				(20,85.95)
                    (25,86.04)
                    (30,86.44)
			};
			\addlegendentry{LiDAR \textit{Rank1}}%第一条线标签

			\addplot+[dashed,semithick,mark=triangle,mark options={scale=2}, color=color1] plot coordinates { 
				(1,52.29)
                    (3,87.66) 
				(5,92.40) 
				(10,94.89)
				(15,95.53)
				(20,95.80)
                    (25,95.74)
                    (30,95.85)
			};
			\addlegendentry{LiDAR \textit{Rank5}}%第一条线标签

			%画第二条线
			\addplot+[semithick, mark=star,mark options={scale=2}, color=color2] plot coordinates {
				(1,4.46)
				(3,26.22)
				(5,42.70)
				(10,60.53)
				(15,66.92)
				(20,69.88)
                    (25,71.69)
                    (30,72.84)
			};
			\addlegendentry{Camera \textit{Rank1}} %第二条线标签
   
			\addplot+[dashed, semithick, mark=star,mark options={scale=2}, color=color2] plot coordinates {
				(1,11.59)
				(3,48.54)
				(5,67.02)
				(10,80.77)
				(15,84.43)
				(20,86.24)
                    (25,87.17)
                    (30,87.74)
			};
			\addlegendentry{Camera \textit{Rank5}} %第二条线标签
   			%画第三条线
			% \addplot+[dashed,semithick,mark options={scale=0}, color=color2] plot coordinates {
			% 	(0,65.85)
			% 	(50,65.85)
			% };
			% \addlegendentry{All Silhouettes} %第三条线标签

			\end{axis}
			\end{tikzpicture}
		}
		\caption{The performance comparison between LiDAR and camera on used frame number in inference.} % 设置caption
		\label{fig:label}  % 设置用于reference的label
	\end{figure}

\section{Qualitative results}
\begin{figure}[ht]
     \centering
     \begin{subfigure}[b]{0.45\textwidth}
         \centering
         \includegraphics[width=0.7\textwidth]{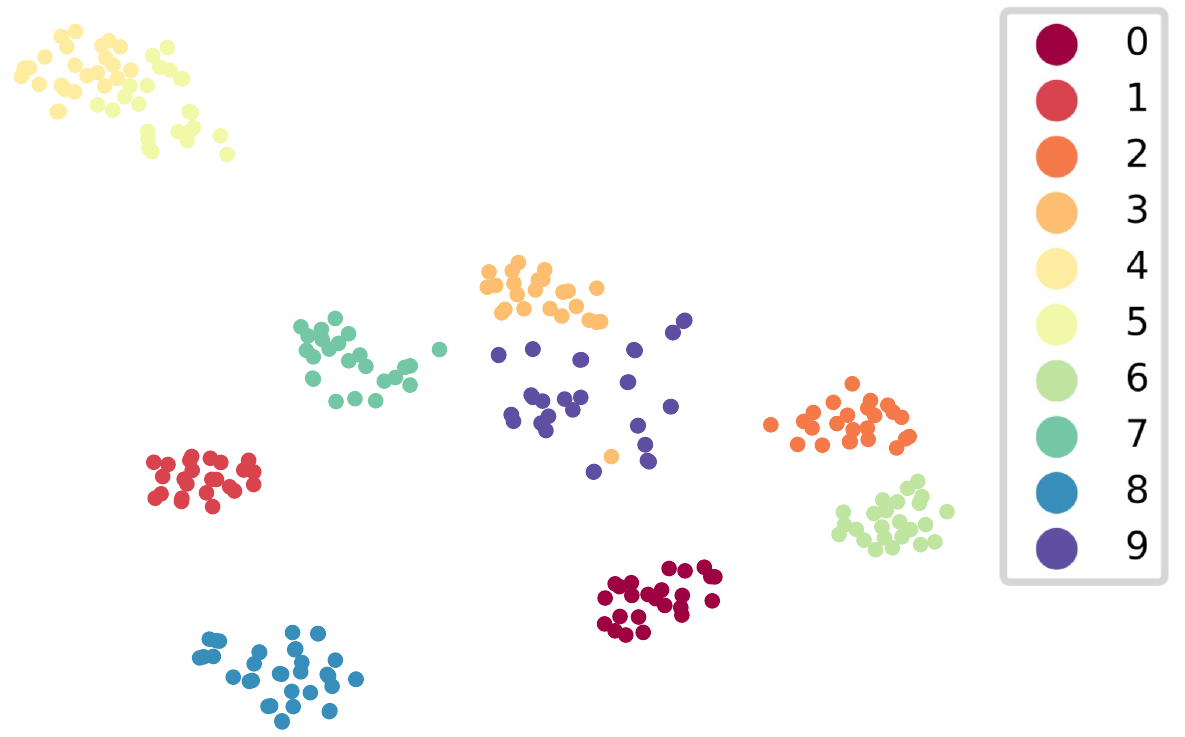}
         \caption{PointNet}
         \label{fig:pointnettsne}
     \end{subfigure}
     \hfill
     \begin{subfigure}[b]{0.45\textwidth}
         \centering
         \includegraphics[width=0.7\textwidth]{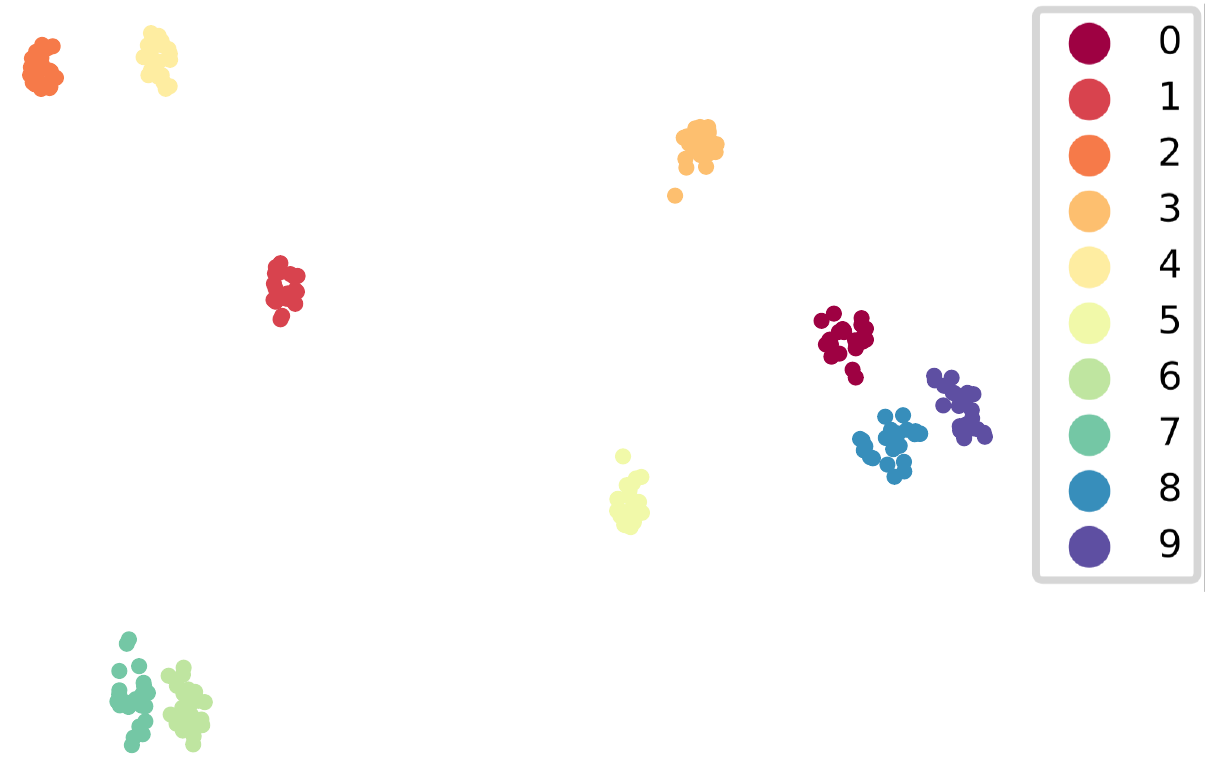}
         \caption{Our LidarGait}
         \label{fig:lidargaittsne}
     \end{subfigure}
        \caption{Feature distributions are visualized by t-SNE. }
        \label{fig:tsne}
\end{figure}

To analyze the performance gap between our LidarGait and representative PointNet, We visualize the feature distribution on the SUSTech1K dataset. We can observe that LidarGait can capture features with clear discrimination. As shown in Fig.~\ref{fig:lidargaittsne}, LidarGait prominently learns the inter-class margin and makes the intra-class distribution more compact. However, the representative point-wise model, PointNet, can only obtain global features as shown in Fig.~\ref{fig:pointnettsne}. PointNet captures features with less discrimination, and its intra-class features distribute sparsely.

\section{Evaluation if Variants as Gallery}
As reported in Tab.~\ref{tab:evaluation}, we evaluate when sequences in variation conditions are gallery sets with normal conditions as probe sets. We observe the performance degradation when normal cases are in the role of probes.

\begin{table}[ht]
\centering
\caption{Performance comparison of different evaluation protocals.\label{tab:evaluation} }

\scalebox{0.9}{
\begin{tabular}{|l|cc|cc|}
\hline
\multicolumn{1}{|c|}{\multirow{2}{*}{\begin{tabular}[c]{@{}c@{}}Evaluation \\ Protocol\end{tabular}}} & \multicolumn{2}{l|}{\begin{tabular}[c]{@{}l@{}}Gallery: \textit{normal}\\ Probe: \textit{\textbf{variation}}\end{tabular}} & \multicolumn{2}{l|}{\begin{tabular}[c]{@{}l@{}}Gallery: \textit{\textbf{variation}}\\ Probe: \textit{normal}\end{tabular}} \\ \cline{2-5} 
\multicolumn{1}{|c|}{}                                                                                & \multicolumn{1}{l|}{\textit{Rank1}}                                & \textit{Rank5}                               & \multicolumn{1}{l|}{\textit{Rank1}}                               & \textit{Rank5}                              \\ \hline
Camera                                                                                                & \multicolumn{1}{l|}{76.12}                                & 89.39                               & \multicolumn{1}{l|}{74.84}                               & 89.28                              \\ \hline
LiDAR                                                                                                 & \multicolumn{1}{l|}{86.77}                                & 96.08                               & \multicolumn{1}{l|}{84.31}                               & 94.84                              \\ \hline
\end{tabular}
}
\vspace{-1mm}
\end{table}

\section{Exemplar Sequences of SUSTech1K}
\begin{figure*}[ht] %插入图片
\centering %图片居中
\includegraphics[width=0.95\linewidth]{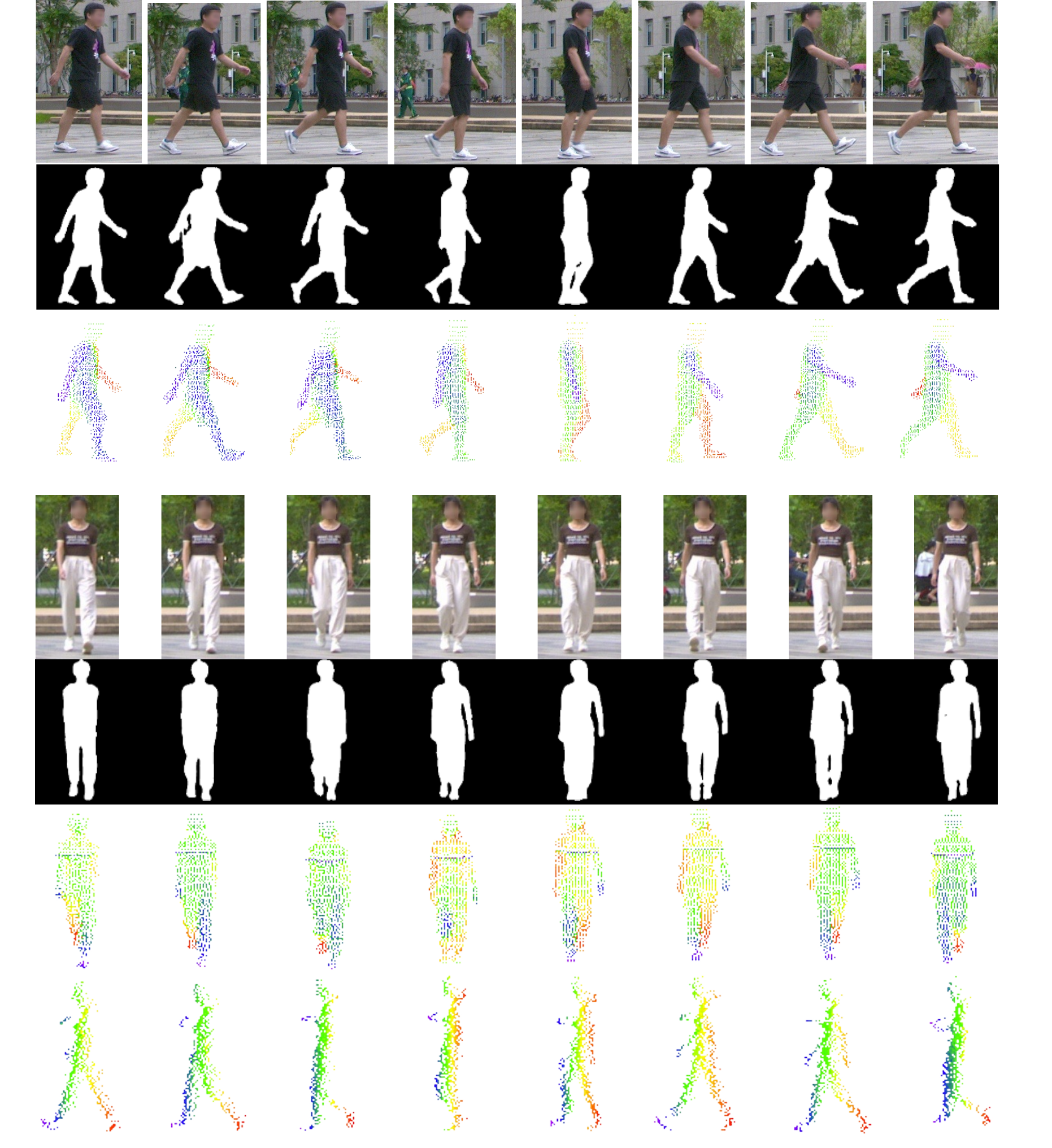}
\caption{Exemplar sequences of SUSTech1K dataset under normal conditions. Eight frames in three modalities are visualized. The top three rows show three gait representations in RGB images, silhouettes, and point clouds. The bottom four rows represent gait representations in RGB images, silhouettes, front-view point clouds, and side-view point clouds. It shows LiDAR provides informative 3D geometry. (Best viewed in color.)}
\label{fig:nm-demo}
\end{figure*}

\begin{figure*}[ht] %插入图片
\centering %图片居中
\includegraphics[width=0.95\linewidth]{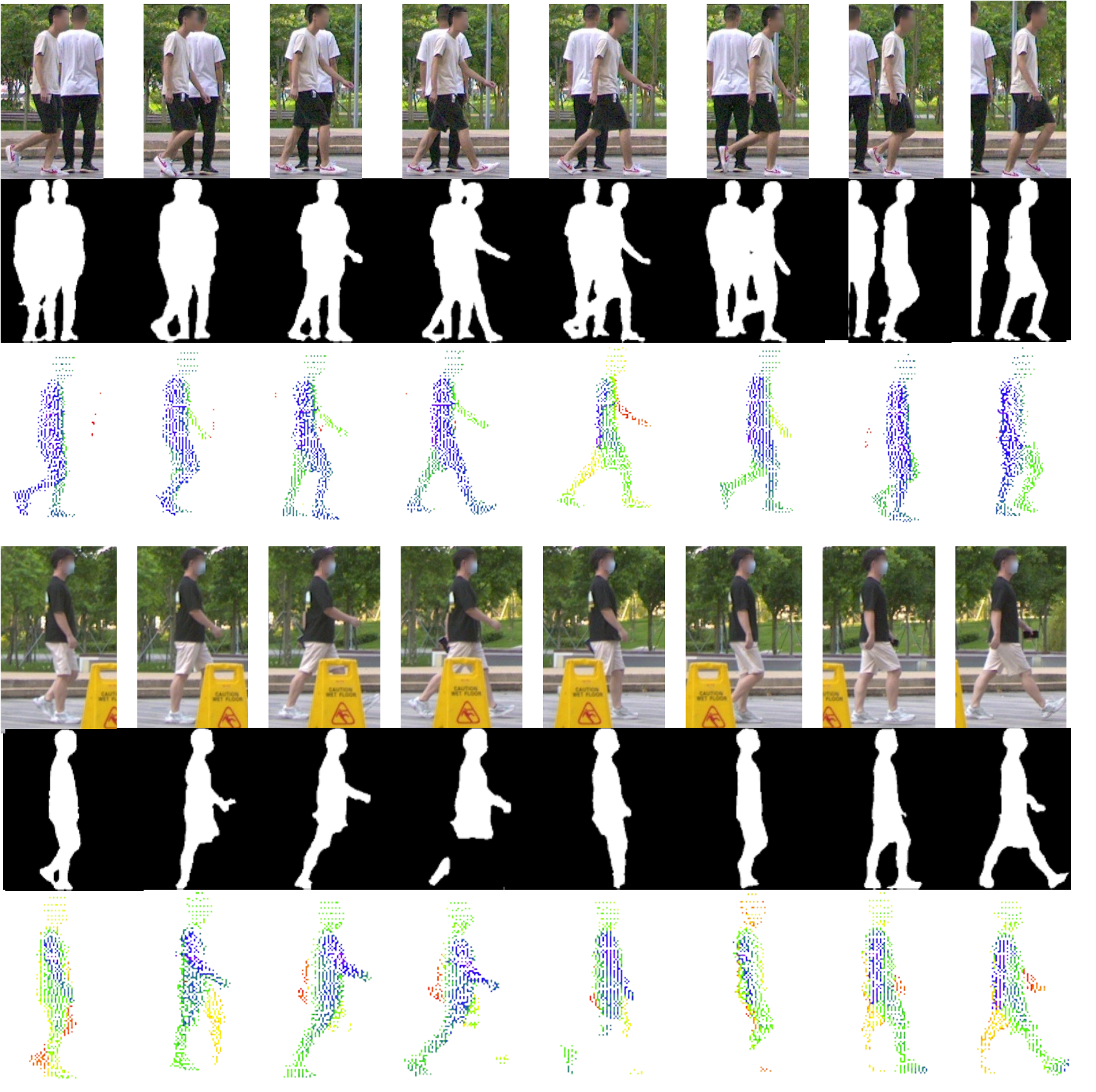}
\caption{Exemplar sequences of SUSTech1K dataset under occlusions. The top three rows show gait representations when another subject overlaps the of-the-interest pedestrian. The bottom three rows show gait representations occluded by a static obstruction. It indicates that LiDAR can provide robust gait representations under occlusion conditions. (Best viewed in color.)}
\label{fig:oc-demo}
\end{figure*}

\begin{figure*}[ht] %插入图片
\centering %图片居中
\includegraphics[width=0.95\linewidth]{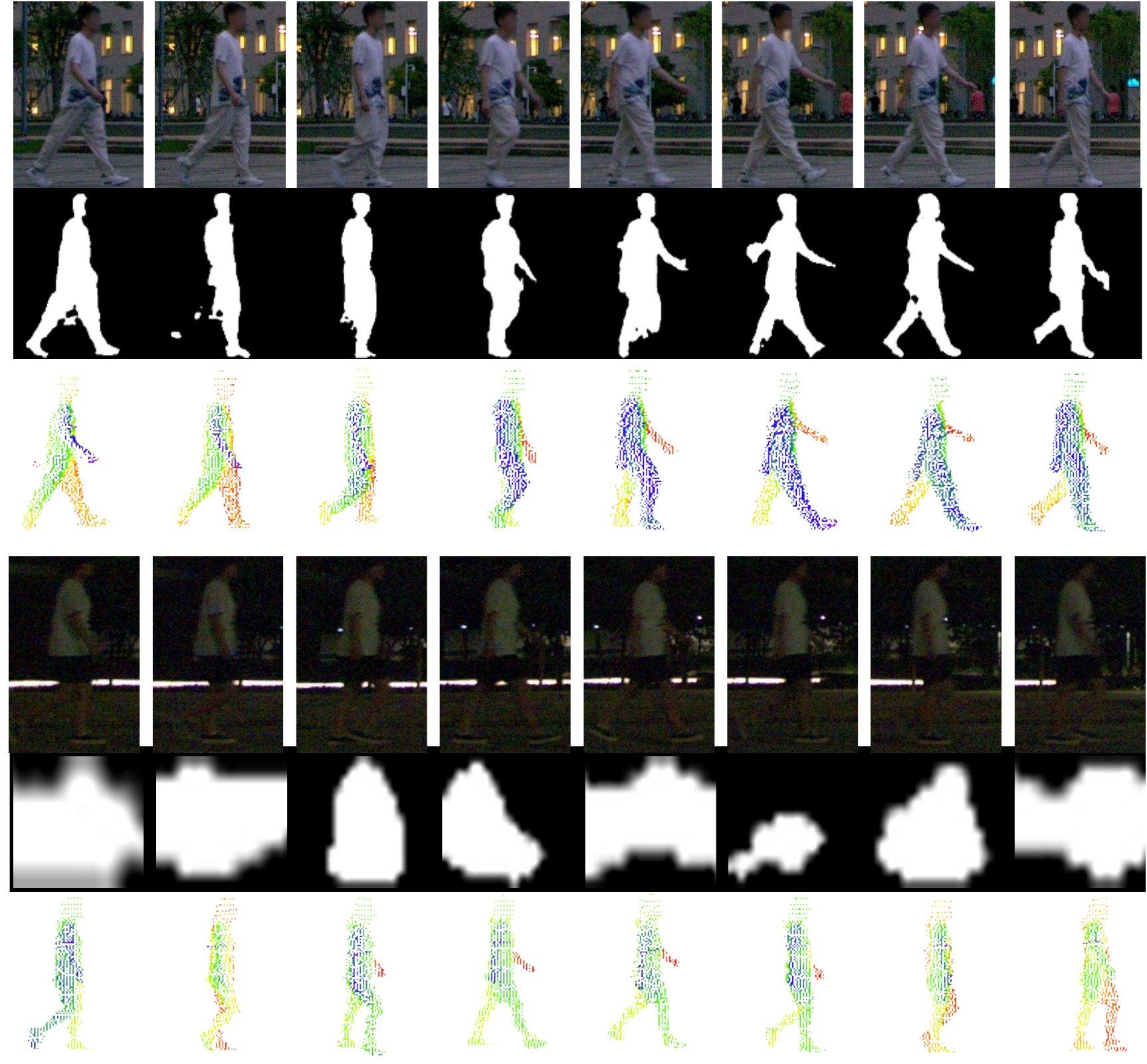}
\caption{Exemplar sequences of SUSTech1K dataset under poor illumination. When illumination is extremely low, human segmentation is barely performed. In contrast, LiDAR provides robust gait representation with point clouds regardless of lighting. (Best viewed in color.)}
\label{fig:nt-demo}
\end{figure*}

To demonstrate the necessity of the SUSTech1K dataset, We show several exemplar sequences of the SUSTech1K dataset under normal, occlusion, and poor illumination conditions in Fig.~\ref{fig:nm-demo}~-~\ref{fig:nt-demo}.

Fig.~\ref{fig:nm-demo} shows that LiDAR provides informative geometry as significant cues that extend gait recognition from 2D to 3D space. The most considerable advantage of LiDAR for gait recognition is that it allows for perspective from another viewpoint, as shown in the bottom row of Fig.~\ref{fig:nm-demo}.

When the pedestrians are occluded, as shown in Fig.~\ref{fig:oc-demo}, the silhouettes obtained by segmentation methods are typical with lower quality. The conventional segmentation methods are based on 2D cameras, but humans live in 3D space, making it difficult to separate the off-the-interest pedestrian from 2D space. LiDAR with precise 3D information can obtain high-quality gait representation under the condition of occlusion.

When the pedestrians are occluded, as shown in Fig.~\ref{fig:oc-demo}, the silhouettes obtained by segmentation methods are typical with lower quality. The conventional segmentation methods are based on 2D cameras, but humans live in 3D space, making it difficult to separate the of-the-interest pedestrian from 2D space. With precise 3D information, LiDAR can obtain high-quality gait representation under occlusion.

\begin{figure*}[ht] %插入图片
\centering %图片居中
     \begin{subfigure}[b]{1
     \textwidth}
         \centering
         \includegraphics[width=1\textwidth]{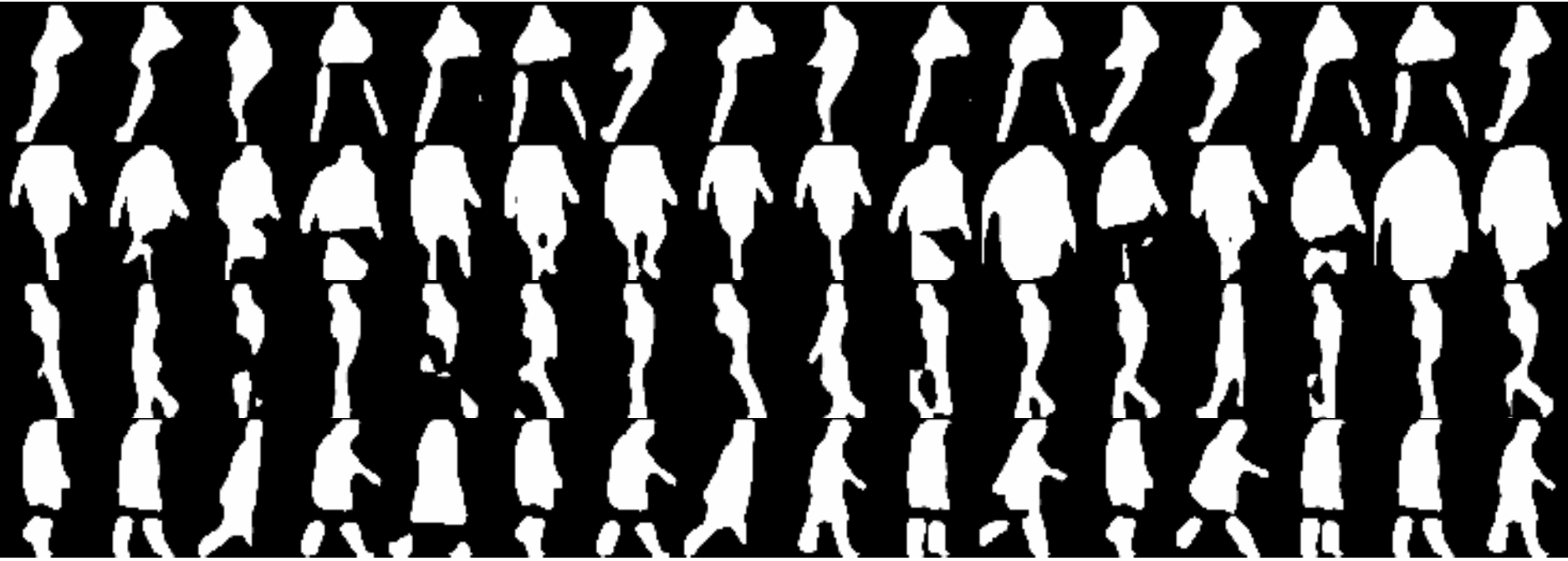}
         \caption{GREW}
         \label{fig:demo-grew}
     \end{subfigure}
     \hfill
     \begin{subfigure}[b]{1\textwidth}
         \centering
         \includegraphics[width=1\textwidth]{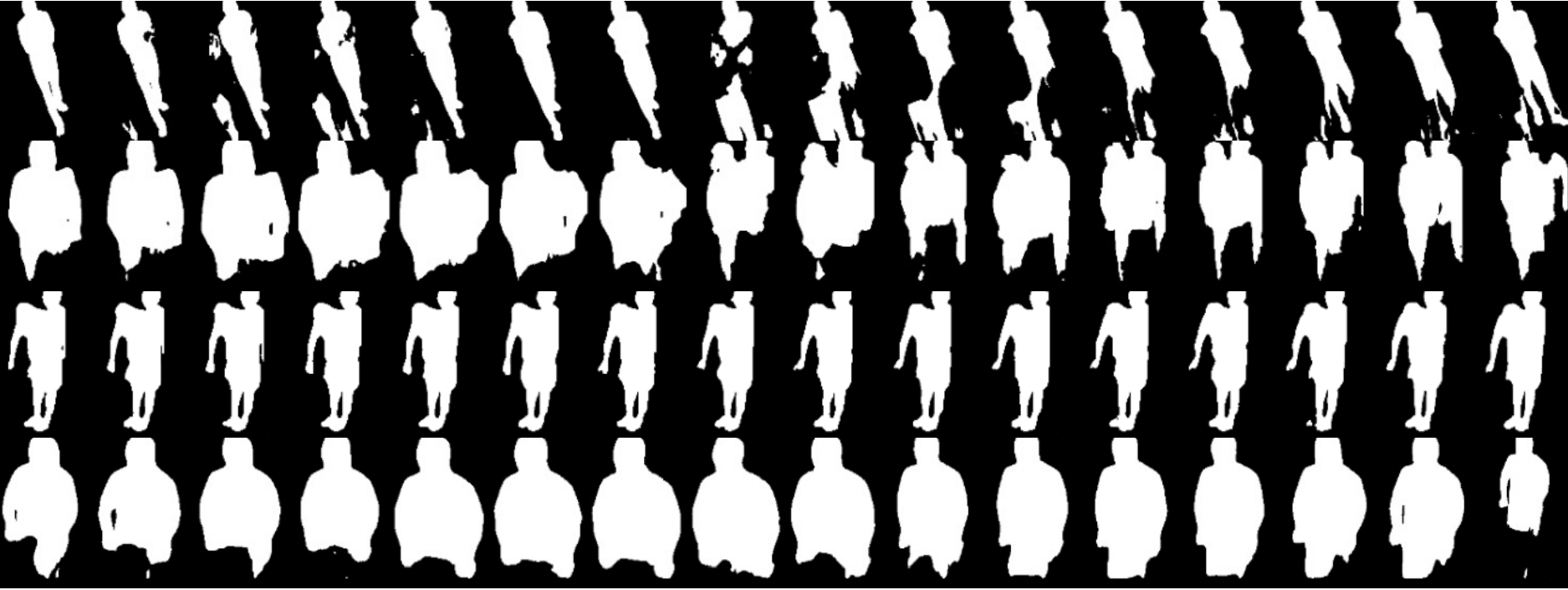}
         \caption{Gait3D}
         \label{fig:demo-gait3d}
     \end{subfigure}
        \caption{Failure silhouettes in the existing in-the-wild datasets. The existing in-the-wild datasets face the issues that segmentation methods fail to provide gait representations with high quality by the effect of many real-world factors.}
        \label{fig:demo-others}
\end{figure*}

In Fig.~\ref{fig:demo-others}, we show gait representations in existing in-the-wild datasets, GREW and Gait3D. We can observe that failure gait representations commonly exist because of various factors in the real world. Therefore, it is necessary to investigate a new way to obtain robust gait representation in such complex scenarios.

%%%%%%%%% REFERENCES
{\small
\bibliographystyle{ieee_fullname}
\bibliography{egbib}

\begin{thebibliography}{10}\itemsep=-1pt

\bibitem{lidarmethod}
Jeongho Ahn, Kazuto Nakashima, Koki Yoshino, Yumi Iwashita, and Ryo Kurazume.
\newblock 2v-gait: Gait recognition using 3d lidar robust to changes in walking
  direction and measurement distance.
\newblock In {\em 2022 IEEE/SICE International Symposium on System Integration
  (SII)}, pages 602--607, 2022.

\bibitem{tunnel}
Gunawan Ariyanto and Mark~S Nixon.
\newblock Model-based 3d gait biometrics.
\newblock In {\em 2011 international joint conference on biometrics (IJCB)},
  pages 1--7, 2011.

\bibitem{lidar2}
Csaba Benedek, Bence G{\'a}lai, Bal{\'a}zs Nagy, and Zsolt Jank{\'o}.
\newblock Lidar-based gait analysis and activity recognition in a 4d
  surveillance system.
\newblock {\em IEEE Transactions on Circuits and Systems for Video Technology},
  28(1):101--113, 2016.

\bibitem{MEI}
A.F. Bobick and J.W. Davis.
\newblock The recognition of human movement using temporal templates.
\newblock {\em IEEE TPAMI}, 23(3):257--267, 2001.

\bibitem{gaitset}
Hanqing Chao, Yiwei He, Junping Zhang, and Jianfeng Feng.
\newblock Gaitset: Regarding gait as a set for cross-view gait recognition.
\newblock In {\em AAAI}, pages 8126--8133, 2019.

\bibitem{stcrowd}
Peishan Cong, Xinge Zhu, Feng Qiao, Yiming Ren, Xidong Peng, Yuenan Hou, Lan
  Xu, Ruigang Yang, Dinesh Manocha, and Yuexin Ma.
\newblock Stcrowd: A multimodal dataset for pedestrian perception in crowded
  scenes.
\newblock In {\em CVPR}, pages 19608--19617, 2022.

\bibitem{doumetagait}
Huanzhang Dou, Pengyi Zhang, and Wei Su.
\newblock Metagait: Learning to learn an omni sample adaptive representation
  for gait recognition.
\newblock In {\em ECCV}, 2022.

\bibitem{dou2021versatilegait}
Huanzhang Dou, Wenhu Zhang, Pengyi Zhang, Yuhan Zhao, Songyuan Li, Zequn Qin,
  Fei Wu, Lin Dong, and Xi Li.
\newblock Versatilegait: A large-scale synthetic gait dataset with
  fine-grainedattributes and complicated scenarios.
\newblock {\em arXiv preprint arXiv:2101.01394}, 2021.

\bibitem{dbscan}
Martin Ester, Hans-Peter Kriegel, J{\"o}rg Sander, Xiaowei Xu, et~al.
\newblock A density-based algorithm for discovering clusters in large spatial
  databases with noise.
\newblock In {\em KDD}, pages 226--231, 1996.

\bibitem{opengait}
Chao Fan, Junhao Liang, Chuanfu Shen, Saihui Hou, Yongzhen Huang, and Shiqi Yu.
\newblock Opengait: Revisiting gait recognition toward better practicality.
\newblock {\em arXiv preprint arXiv:2211.06597}, 2022.

\bibitem{gaitpart}
Chao Fan, Yunjie Peng, Chunshui Cao, Xu Liu, Saihui Hou, Jiannan Chi, Yongzhen
  Huang, Qing Li, and Zhiqiang He.
\newblock Gaitpart: Temporal part-based model for gait recognition.
\newblock In {\em CVPR}, pages 14225--14233, 2020.

\bibitem{kinect}
P{\'e}ter Fankhauser, Michael Bloesch, Diego Rodriguez, Ralf Kaestner, Marco
  Hutter, and Roland Siegwart.
\newblock Kinect v2 for mobile robot navigation: Evaluation and modeling.
\newblock In {\em 2015 International Conference on Advanced Robotics (ICAR)},
  pages 388--394, 2015.

\bibitem{ge2021yolox}
Zheng Ge, Songtao Liu, Feng Wang, Zeming Li, and Jian Sun.
\newblock Yolox: Exceeding yolo series in 2021.
\newblock {\em arXiv preprint arXiv:2107.08430}, 2021.

\bibitem{kitti}
Andreas Geiger, Philip Lenz, Christoph Stiller, and Raquel Urtasun.
\newblock Vision meets robotics: The kitti dataset.
\newblock {\em The International Journal of Robotics Research},
  32(11):1231--1237, 2013.

\bibitem{simpleview}
Ankit Goyal, Hei Law, Bowei Liu, Alejandro Newell, and Jia Deng.
\newblock Revisiting point cloud shape classification with a simple and
  effective baseline.
\newblock In {\em ICML}, pages 3809--3820, 2021.

\bibitem{gei}
Jinguang Han and Bir Bhanu.
\newblock Individual recognition using gait energy image.
\newblock {\em IEEE TPAMI}, 28(2):316--322, 2005.

\bibitem{tumgait}
Martin Hofmann, J{\"u}rgen Geiger, Sebastian Bachmann, Bj{\"o}rn Schuller, and
  Gerhard Rigoll.
\newblock The tum gait from audio, image and depth (gaid) database: Multimodal
  recognition of subjects and traits.
\newblock {\em Journal of Visual Communication and Image Representation},
  25(1):195--206, 2014.

\bibitem{cstl}
Xiaohu Huang, Duowang Zhu, Hao Wang, Xinggang Wang, Bo Yang, Botao He, Wenyu
  Liu, and Bin Feng.
\newblock Context-sensitive temporal feature learning for gait recognition.
\newblock In {\em ICCV}, pages 12909--12918, 2021.

\bibitem{oulp}
H. Iwama, M. Okumura, Y. Makihara, and Y. Yagi.
\newblock The ou-isir gait database comprising the large population dataset and
  performance evaluation of gait recognition.
\newblock {\em IEEE TIFS}, 7, Issue 5:1511--1521, 2012.

\bibitem{ky4d}
Y. Iwashita, R. Baba, K. Ogawara, and R. Kurazume.
\newblock Person identification from spatio-temporal 3d gait.
\newblock In {\em Int. Conf. Emerging Security Technologies (EST)}, 2010.

\bibitem{groundremoval}
Seungjae Lee, Hyungtae Lim, and Hyun Myung.
\newblock Patchwork++: Fast and robust ground segmentation solving partial
  under-segmentation using 3d point cloud.
\newblock In {\em 2022 IEEE/RSJ International Conference on Intelligent Robots
  and Systems (IROS)}, pages 13276--13283, 2022.

\bibitem{point2depth}
Bo Li, Tianlei Zhang, and Tian Xia.
\newblock Vehicle detection from 3d lidar using fully convolutional network.
\newblock {\em arXiv preprint arXiv:1608.07916}, 2016.

\bibitem{lixiangpose}
Xiang Li, Yasushi Makihara, Chi Xu, and Yasushi Yagi.
\newblock End-to-end model-based gait recognition using synchronized multi-view
  pose constraint.
\newblock In {\em ICCVW}, pages 4106--4115, 2021.

\bibitem{oumvlpmesh}
Xiang Li, Yasushi Makihara, Chi Xu, and Yasushi Yagi.
\newblock Multi-view large population gait database with human meshes and its
  performance evaluation.
\newblock {\em IEEE TBIOM}, 4(2):234--248, 2022.

\bibitem{lixiangpose2}
Xiang Li, Yasushi Makihara, Chi Xu, Yasushi Yagi, Shiqi Yu, and Mingwu Ren.
\newblock End-to-end model-based gait recognition.
\newblock In {\em ACCV}, 2020.

\bibitem{liang2022gaitedge}
Junhao Liang, Chao Fan, Saihui Hou, Chuanfu Shen, Yongzhen Huang, and Shiqi Yu.
\newblock Gaitedge: Beyond plain end-to-end gait recognition for better
  practicality.
\newblock {\em arXiv preprint arXiv:2203.03972}, 2022.

\bibitem{posegait}
Rijun Liao, Shiqi Yu, Weizhi An, and Yongzhen Huang.
\newblock A model-based gait recognition method with body pose and human prior
  knowledge.
\newblock {\em PR}, 98:107069, 2020.

\bibitem{lin2022uncertainty}
Beibei Lin, Chen Liu, Lincheng Li, Robby~T Tan, and Xin Yu.
\newblock Uncertainty-aware gait recognition via learning from dirichlet
  distribution-based evidence.
\newblock {\em arXiv preprint arXiv:2211.08007}, 2022.

\bibitem{lin2021gaitmask}
Beibei Lin, Yu Liu, and Shunli Zhang.
\newblock Gaitmask: Mask-based model for gait recognition.
\newblock In {\em BMVC}, pages 1--12, 2021.

\bibitem{mt3d}
Beibei Lin, Shunli Zhang, and Feng Bao.
\newblock Gait recognition with multiple-temporal-scale 3d convolutional neural
  network.
\newblock In {\em Proceedings of the 28th ACM international conference on
  multimedia}, pages 3054--3062, 2020.

\bibitem{gaitgl}
Beibei Lin, Shunli Zhang, and Xin Yu.
\newblock Gait recognition via effective global-local feature representation
  and local temporal aggregation.
\newblock In {\em ICCV}, pages 14648--14656, 2021.

\bibitem{liu2021paddleseg}
Yi Liu, Lutao Chu, Guowei Chen, Zewu Wu, Zeyu Chen, Baohua Lai, and Yuying Hao.
\newblock Paddleseg: A high-efficient development toolkit for image
  segmentation.
\newblock {\em arXiv preprint arXiv:2101.06175}, 2021.

\bibitem{waymo}
Jieru Mei, Alex~Zihao Zhu, Xinchen Yan, Hang Yan, Siyuan Qiao, Liang-Chieh
  Chen, and Henrik Kretzschmar.
\newblock Waymo open dataset: Panoramic video panoptic segmentation.
\newblock In {\em ECCV}, pages 53--72, 2022.

\bibitem{resgait}
Zihao Mu, Francisco~M Castro, Manuel~J Mar{\'\i}n-Jim{\'e}nez, Nicol{\'a}s
  Guil, Yan-Ran Li, and Shiqi Yu.
\newblock Resgait: The real-scene gait dataset.
\newblock In {\em 2021 IEEE International Joint Conference on Biometrics
  (IJCB)}, pages 1--8, 2021.

\bibitem{nixon2010book}
Mark~S Nixon, Tieniu Tan, and Rama Chellappa.
\newblock {\em Human identification based on gait}, volume~4.
\newblock Springer Science \& Business Media, 2010.

\bibitem{pointnet}
Charles~R Qi, Hao Su, Kaichun Mo, and Leonidas~J Guibas.
\newblock Pointnet: Deep learning on point sets for 3d classification and
  segmentation.
\newblock In {\em CVPR}, pages 652--660, 2017.

\bibitem{pointnet++}
Charles~Ruizhongtai Qi, Li Yi, Hao Su, and Leonidas~J Guibas.
\newblock Pointnet++: Deep hierarchical feature learning on point sets in a
  metric space.
\newblock {\em NeurIPS}, 30, 2017.

\bibitem{reversemask}
Chuanfu Shen, Beibei Lin, Shunli Zhang, George~Q Huang, Shiqi Yu, and Xin Yu.
\newblock Gait recognition with mask-based regularization.
\newblock {\em arXiv preprint arXiv:2203.04038}, 2022.

\bibitem{survey}
Chuanfu Shen, Shiqi Yu, Jilong Wang, George~Q Huang, and Liang Wang.
\newblock A comprehensive survey on deep gait recognition: algorithms, datasets
  and challenges.
\newblock {\em arXiv preprint arXiv:2206.13732}, 2022.

\bibitem{casiae}
Chunfeng Song, Yongzhen Huang, Weining Wang, and Liang Wang.
\newblock Casia-e: a large comprehensive dataset for gait recognition.
\newblock {\em IEEE Transactions on Pattern Analysis and Machine Intelligence},
  2022.

\bibitem{mvcnn}
Hang Su, Subhransu Maji, Evangelos Kalogerakis, and Erik Learned-Miller.
\newblock Multi-view convolutional neural networks for 3d shape recognition.
\newblock In {\em ICCV}, pages 945--953, 2015.

\bibitem{oumvlp}
Noriko Takemura, Yasushi Makihara, Daigo Muramatsu, Tomio Echigo, and Yasushi
  Yagi.
\newblock Multi-view large population gait dataset and its performance
  evaluation for cross-view gait recognition.
\newblock {\em IPSJ Trans. on Computer Vision and Applications}, 10(4):1--14,
  2018.

\bibitem{nightgait}
Daoliang Tan, Kaiqi Huang, Shiqi Yu, and Tieniu Tan.
\newblock Efficient night gait recognition based on template matching.
\newblock In {\em ICPR}, pages 1000--1003, 2006.

\bibitem{gaitgraph}
Torben Teepe, Ali Khan, Johannes Gilg, Fabian Herzog, Stefan H\"ormann, and
  Gerhard Rigoll.
\newblock Gait{G}raph: Graph convolutional network for skeleton-based gait
  recognition.
\newblock In {\em ICIP}, pages 2314--2318, 2021.

\bibitem{wang2010chrono}
Chen Wang, Junping Zhang, Jian Pu, Xiaoru Yuan, and Liang Wang.
\newblock Chrono-gait image: A novel temporal template for gait recognition.
\newblock In {\em ECCV}, pages 257--270, 2010.

\bibitem{casiaa}
Liang Wang, Tieniu Tan, Huazhong Ning, and Weiming Hu.
\newblock Silhouette analysis-based gait recognition for human identification.
\newblock {\em IEEE TPAMI}, 25(12):1505--1518, 2003.

\bibitem{DGCNN}
Yue Wang, Yongbin Sun, Ziwei Liu, Sanjay~E Sarma, Michael~M Bronstein, and
  Justin~M Solomon.
\newblock Dynamic graph cnn for learning on point clouds.
\newblock {\em Acm Transactions On Graphics (tog)}, 38(5):1--12, 2019.

\bibitem{viewgcn}
Xin Wei, Ruixuan Yu, and Jian Sun.
\newblock View-gcn: View-based graph convolutional network for 3d shape
  analysis.
\newblock In {\em CVPR}, pages 1850--1859, 2020.

\bibitem{lidar1}
Hiroyuki Yamada, Jeongho Ahn, Oscar~Martinez Mozos, Yumi Iwashita, and Ryo
  Kurazume.
\newblock Gait-based person identification using 3d lidar and long short-term
  memory deep networks.
\newblock {\em Advanced Robotics}, 34(18):1201--1211, 2020.

\bibitem{casiab}
Shiqi Yu, Daoliang Tan, and Tieniu Tan.
\newblock A framework for evaluating the effect of view angle, clothing and
  carrying condition on gait recognition.
\newblock In {\em ICPR}, pages 441--444, 2006.

\bibitem{zhang2022bytetrack}
Yifu Zhang, Peize Sun, Yi Jiang, Dongdong Yu, Fucheng Weng, Zehuan Yuan, Ping
  Luo, Wenyu Liu, and Xinggang Wang.
\newblock Bytetrack: Multi-object tracking by associating every detection box.
\newblock In {\em ECCV}, pages 1--21, 2022.

\bibitem{fvg}
Ziyuan Zhang, Luan Tran, Feng Liu, and Xiaoming Liu.
\newblock On learning disentangled representations for gait recognition.
\newblock {\em IEEE TPAMI}, 2020.

\bibitem{gaitnet}
Ziyuan Zhang, Luan Tran, Xi Yin, Yousef Atoum, Xiaoming Liu, Jian Wan, and
  Nanxin Wang.
\newblock Gait recognition via disentangled representation learning.
\newblock In {\em CVPR}, pages 4710--4719, 2019.

\bibitem{pointtransformer}
Hengshuang Zhao, Li Jiang, Jiaya Jia, Philip~HS Torr, and Vladlen Koltun.
\newblock Point transformer.
\newblock In {\em Proceedings of the IEEE/CVF International Conference on
  Computer Vision}, pages 16259--16268, 2021.

\bibitem{hopgait}
Jinkai Zheng, Xinchen Liu, Xiaoyan Gu, Yaoqi Sun, Chuang Gan, Jiyong Zhang, Wu
  Liu, and Chenggang Yan.
\newblock Gait recognition in the wild with multi-hop temporal switch.
\newblock In {\em ACMMM}, pages 6136--6145, 2022.

\bibitem{gait3d}
Jinkai Zheng, Xinchen Liu, Wu Liu, Lingxiao He, Chenggang Yan, and Tao Mei.
\newblock Gait recognition in the wild with dense 3d representations and a
  benchmark.
\newblock In {\em ICCV}, pages 20228--20237, 2022.

\bibitem{3dreid}
Zhedong Zheng, Xiaohan Wang, Nenggan Zheng, and Yi Yang.
\newblock Parameter-efficient person re-identification in the 3d space.
\newblock {\em IEEE TNNLS}, 2022.

\bibitem{grew}
Zheng Zhu, Xianda Guo, Tian Yang, Junjie Huang, Jiankang Deng, Guan Huang,
  Dalong Du, Jiwen Lu, and Jie Zhou.
\newblock Gait recognition in the wild: A benchmark.
\newblock In {\em CVPR}, pages 14789--14799, 2021.

\end{thebibliography}
}

\end{document}